\documentclass[preprint]{article}

\PassOptionsToPackage{numbers, compress}{natbib}

\usepackage{neurips_2026}

\usepackage[utf8]{inputenc} 
\usepackage[T1]{fontenc}    
\usepackage{hyperref}       
\usepackage{url}            
\usepackage{booktabs}       
\usepackage{amsfonts}       
\usepackage{nicefrac}       
\usepackage{microtype}      
\usepackage{xcolor}         

\usepackage{amsmath}
\usepackage{amssymb}
\usepackage{graphicx}
\usepackage{algorithm}
\usepackage{algpseudocode}
\usepackage{multirow}
\usepackage{wrapfig}
\usepackage{cancel}
\usepackage{flafter}
\usepackage{placeins}
\usepackage{subcaption}

\title{Emergent Communication between Heterogeneous Visual Agents through Decentralized Learning}

\author{%
  Mikako Ochiai \quad Masatoshi Nagano \quad Tadahiro Taniguchi \\
  Graduate School of Informatics, Kyoto University, Kyoto 606-8501, Japan \\
  \texttt{ochiai.mikako.36x@st.kyoto-u.ac.jp} \quad
  \texttt{nagano.masatoshi.5e@kyoto-u.ac.jp} \\
  \texttt{taniguchi@i.kyoto-u.ac.jp}
}

\begin{document}

\maketitle

\begin{abstract}
Symbols are shared, but perception is private. We study emergent communication between heterogeneous visual agents through decentralized learning, asking what visual information can become shareable when agents have different visual representations. Instead of optimizing messages through a shared external communicative objective, our agents exchange only discrete token sequences and update their own models using local perceptual evidence. This setting focuses on an underexplored aspect of emergent communication, examining whether common symbols can arise without shared perceptual access, and how the similarity between private visual spaces constrains the content and symmetry of the resulting language. We instantiate this setting in the Metropolis-Hastings Captioning Game (MHCG), where two agents collaboratively form shared captions by exchanging proposed token sequences that a listener accepts or rejects using an MH-style criterion evaluated against its own visual features. We compare three pairings of frozen visual encoders, with agents starting from randomly initialized text modules. Experiments on MS-COCO show that MHCG produces visually informative shared token sequences that outperform a no-communication baseline in cross-agent alignment, visual-feature prediction, and image-text retrieval; all cross-agent metrics decline as encoder mismatch increases. Moderate encoder heterogeneity reduces the number of shared sequences while preserving per-sequence visual specificity, whereas stronger encoder heterogeneity yields fewer, coarser, and more asymmetric sequences. Ablations show that listener-side MH acceptance is critical for avoiding degenerate token formation. These results suggest that shared symbols can arise from local perceptual evaluation alone, with visual representational similarity across encoders shaping both the content and symmetry of the resulting language.
\end{abstract}

\section{Introduction}
\label{sec:intro}
Communication requires agents to coordinate public symbols, but each agent grounds
those symbols in a private perceptual representation that others cannot directly observe\citep{clark1996using}. 
Such visual representations may differ
across architectures, training histories, embodiments, or sensory systems\citep{von2013foray}. 
Public symbols therefore act as an interface between private visual
spaces, raising the question of which aspects of visual representations
can be preserved in symbols that agents come to share.
More specifically, when agents' visual representations differ, what
visual information remains shareable through public symbols, and do the
resulting symbols align with both agents' representational structures or
privilege one agent's visual space \citep{taniguchi2025qualia}?

Emergent communication (EC) is the computational framework most directly
suited to studying such questions, examining how language-like
communication systems arise between interacting agents
\citep{lazaridou2020emergent,peters2025survey}. The dominant paradigm in
EC is referential signaling, in which a speaker's message must enable
a listener to identify a target image, and emergent symbols are shaped
by receiver-success rewards or gradients through relaxed message
channels \citep{lazaridou2018emergence,havrylov2017emergence,
mordatch2018emergence}. Within this paradigm, much work has examined the
\textit{form} of emergent symbols (compositionality, object grounding, sequence
structure), and a more recent line varies visual backbones or
sensory modalities to study representational heterogeneity
\citep{mahaut2025referential,pitzer2026learning}.
 However, the shared
external task signal that anchors this paradigm couples agents through
a common communicative objective, which limits the leverage on our
question because emergent symbols may become useful by being optimized
for the shared signal rather than by being locally grounded in each
agent's private perceptual evidence.
This motivates a decentralized setting in which
agents are coupled only through their exchanged messages and update from their own perceptual evidence \citep{clark1991grounding,taniguchi2024cpc,
taniguchi2025genemcom,taniguchi2023mhng}.

\begin{figure}[t]
  \centering
  \includegraphics[width=\linewidth]{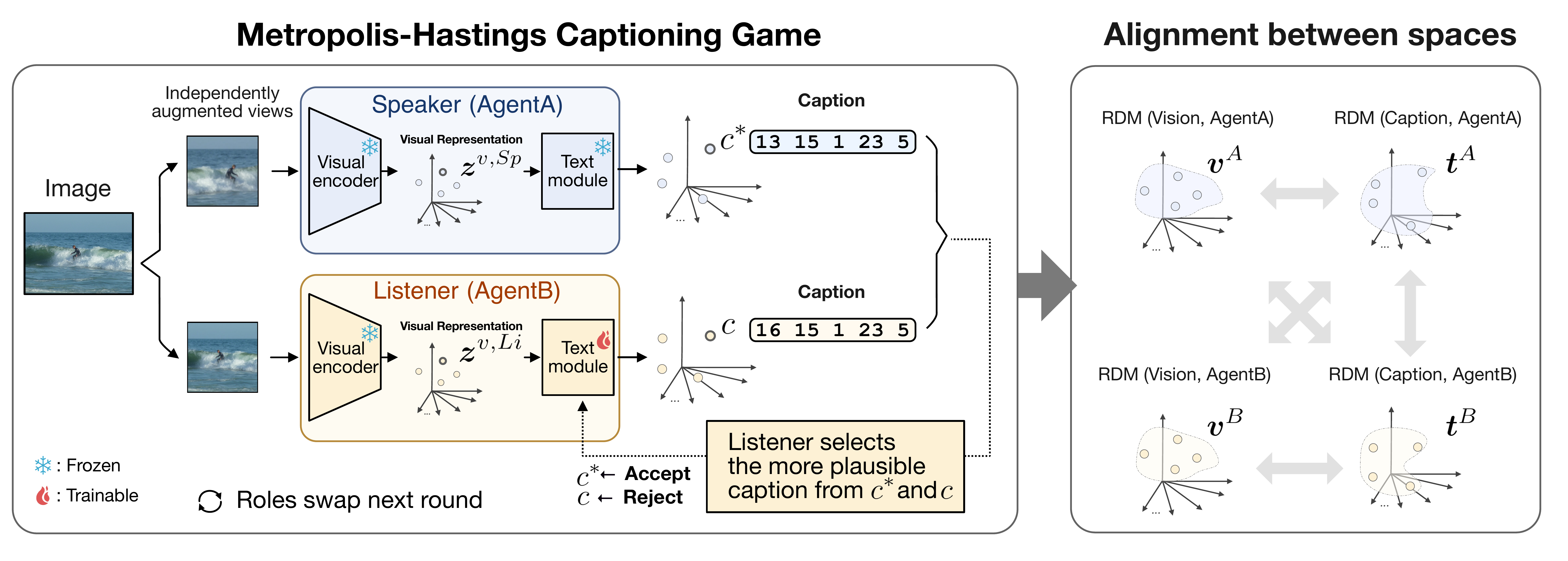}
  \caption{%
  Metropolis--Hastings Captioning Game for heterogeneous visual agents.
Two agents observe independently augmented views of the same image through private frozen visual encoders and exchange only discrete token sequences. The speaker proposes a caption $c^\ast$, and the listener compares it with its own caption $c$, accepting the proposal or retaining its own caption as the learning target according to listener-side visual evidence. The speaker and listener roles swap in the next round. We construct representational dissimilarity matrices (RDMs) of
pairwise distances within visual representations and caption, and
correlate them through the representational similarity analysis (RSA). This
operationalizes the central question of whether public token sequences
preserve information from both private visual spaces, or whether
encoder mismatch makes the emergent symbols align more strongly with
one agent's visual geometry.
}
  \label{fig:task}
\end{figure}

Collective Predictive Coding (CPC) formalizes such a decentralized
setting, modeling symbols as shared latent variables that each agent
infers from its own private observations
\citep{taniguchi2024cpc,taniguchi2025genemcom}. CPC has recently been
derived as a specific extension of free-energy minimization to symbol
emergence \citep{taniguchi2025cpcms}, consistent with extensions of
the free-energy principle \citep{friston2010free} to multi-agent
settings through active inference accounts of dyadic communication
\citep{friston2015active} and federated inference for belief sharing
under a common world model \citep{friston2024federated}. Decentralized inference in the CPC framework can be
approximated by language games based on Metropolis--Hastings (MH)
sampling \citep{taniguchi2023mhng,inukai2023rmhng,hoang2024multimodal,hoang2024compositionality}. In these games agents alternate speaker and listener roles and
the listener accepts or rejects proposed symbols through a
likelihood ratio, approximating posterior inference over the shared latent variable \citep{taniguchi2023mhng}.
Motivated by this line of MH language-game research,
\citet{matsui2025mhcg} introduced the Metropolis--Hastings Captioning
Game (MHCG) for vision--language model (VLM) agents on natural images,
using ProbVLM \citep{upadhyay2023probvlm} to estimate the
likelihood used in the MH acceptance step.

We use MHCG to study how heterogeneity between agents' visual
representations shapes the symbols they come to share. 
Our setting differs from \citet{matsui2025mhcg} in using randomly
initialized text modules and heterogeneous frozen visual encoders.
Thus, the exchanged token sequences can become communicative only through
interaction, rather than by inheriting semantics from pretrained language
modules (Section~\ref{sec:related}). We implement
this setting with two BLIP-style multimodal agents 
\citep{li2022blip} that observe independently augmented views of 
the same image and exchange token sequences in the captioning game (Figure~\ref{fig:task}). We evaluate our system on MS-COCO \citep{lin2014coco} under three
visual-encoder conditions, each defined by a pair of frozen encoders:
one homogeneous condition and two heterogeneous conditions. To connect the
emergent token sequences back to the representational question above, 
we measure whether the two agents produce similar token-sequence
structures and whether those structures align with visual representations using representational similarity analysis (RSA) \citep{kriegeskorte2008representational, sucholutsky2025getting}. 
We additionally evaluate the visual information carried by the token sequences through cross-agent
visual-feature prediction and image-text retrieval.
We further assess whether the MH likelihood-ratio filter provides a useful
acceptance mechanism in this VLM setting by comparing it with listener-side
discriminative matching, rate-matched random acceptance, and the absence of
acceptance.

\paragraph{Contributions.}
\begin{enumerate}
\item We introduce a decentralized VLM language-game setting for asking
what visual representational structure becomes shareable between agents
with heterogeneous visual encoders, without shared communicative
task rewards, cross-agent task losses, cross-agent gradients, or parameter
sharing.

\item We show that MHCG produces shared token sequences that preserve
visual information across agents, outperforming a no-communication
baseline in inter-agent text-text alignment, text-vision alignment, visual-feature
prediction, and image--text retrieval.

\item We characterize how encoder heterogeneity changes the resulting
shared symbols: all cross-agent metrics decrease with encoder
mismatch; moderate mismatch reduces the number of shared sequences while
preserving per-sequence visual specificity; and strong mismatch produces
fewer and coarser sequences with a directional bias toward one agent's
representational structure

\item We show that listener-side MH likelihood-ratio filtering is important
in this VLM instantiation, preventing degenerate token formation and
supporting visual-information transfer compared with discriminative,
random, all-accept, and no-communication controls.

\end{enumerate}

\section{Related Work}
\label{sec:related}

\paragraph{Heterogeneous agents in emergent communication.}
Most EC studies of agent diversity or perceptual heterogeneity still
ground communication in a shared task signal, leaving open how shared
symbols form when agents rely only on private perceptual evidence.
Some EC studies introduce diversity at the population or training level:
\citet{rita2022heterogeneity} studied how population
heterogeneity shapes emergent languages, and
\citet{michel2023revisiting} found that limiting co-adaptation
across sender--receiver pairs improves compositionality.
These works vary population composition or training dynamics while keeping
perception identical across agents.
More directly related to our setting, a separate line of work has begun to vary the perceptual
systems themselves.
\citet{mahaut2025referential} studied referential communication
in communities of heterogeneous pre-trained visual networks, showing that
such networks can learn a shared self-supervised protocol that captures
high-level semantic features.
\citet{pitzer2026learning} studied EC in agents with different
sensory modalities. Unlike these referential-game settings, our agents are not optimized for
receiver success under a shared external task signal. They update only
from evidence evaluated within their own visual representations.
We do not include a referential-game baseline, because standard
referential games optimize messages using receiver success rewards or
task losses, introducing the shared external signal that our decentralized
setting is designed to exclude.


\section{Methods}
\label{sec:methods}


\paragraph{Architecture}
\label{subsec:generative}

Each agent consists of a frozen visual encoder and randomly
initialized text modules connected to a shared
probabilistic embedding space through ProbVLM.
Where \citet{matsui2025mhcg} represent each agent's internal
state as a single latent variable, we decompose it into
modality-specific features ($z^{n,v}$, $z^{n,t}$)
(Appendix~\ref{app:mstep}).
We realize this structure with a BLIP-style multimodal
architecture \citep{li2022blip}
(single-agent internals illustrated in
Figure~\ref{fig:architecture} of Appendix~\ref{app:implementation};
see also Algorithm~\ref{alg:mhcg}).
Each agent $n \in \{A, B\}$ has a frozen pretrained
ViT-based visual encoder $\mathrm{Enc}^{n,v}$
\citep{dosovitskiy2020image}, whose specific instantiation
depends on the encoder condition defined in
Section~\ref{sec:expsetup}. The visual encoder produces visual features
$z^{n,v} = \mathrm{Enc}^{n,v}(o)$ from an input image $o$.
Each agent also has a text encoder/decoder that uses the
BERT-Tiny configuration from the 24 smaller BERT models
released with the google-research BERT repository
\citep{turc2019wellread},
but all text-module weights are initialized from scratch; no
pretrained BERT checkpoint is loaded.
This module is parameterized by $\theta^n$ and shares Feed-Forward and
Self-Attention layers between encoder and decoder modes
\citep{li2022blip}: the encoder maps token sequences $c$ to
text-side features $z^{n,t}$, and the decoder
$\mathrm{Dec}^n$ generates token sequences conditioned on
$z^{n,v}$.
An aggregation function $\psi$ (by default, first-token selection)
followed by a linear projection maps encoder outputs into a shared
latent space:
$\boldsymbol{\mu}^{n,m} = \mathrm{Proj}^{n,m}(\psi(z^{n,m}))$
for modality $m \in \{v, t\}$.
A ProbVLM adapter \citep{upadhyay2023probvlm}
(parameterized by $\phi^{n,m}$) then maps each projected embedding
to diagonal density parameters:
\begin{align}
  \hat{\lambda}^{n,m}
  = \mathrm{ProbVLM}(\boldsymbol{\mu}^{n,m};\, \phi^{n,m}),
  \quad m \in \{v, t\}.
  \label{eq:probvlm}
\end{align}
In the implementation, $\hat{\lambda}^{n,m}$ consists of a
location and scale for a diagonal generalized Gaussian
distribution with shape parameter fixed to $\beta=2$, i.e.,
a diagonal Gaussian up to the ProbVLM scale parameterization.
During MH acceptance, the listener samples a stochastic visual
embedding $\hat{h}^{\mathrm{Li},v}$ from its vision-side density,
while the text-side density parameters
$\hat{\lambda}^{\mathrm{Li},t}(c)$ define the likelihood assigned
to a candidate caption (Eq.~\ref{eq:mh}).

\paragraph{Metropolis--Hastings Captioning Game}
\label{subsec:mhcg}

Following \citet{matsui2025mhcg}, we use a one-sample,
one-step MH-style acceptance filter motivated by decentralized
Bayesian inference over the intractable posterior
$p(c \mid o^A, o^B)$.
The training procedure alternates between caption updates
(MH acceptance, used as an E-step-like update) and parameter
updates (a surrogate M-step); the correspondence between the BLIP
losses and this update is discussed in Appendix~\ref{app:mstep}.

The listener accepts or rejects the speaker's proposed token
sequence through a likelihood ratio evaluated against its own
visual features.
The agents alternate speaker ($\mathrm{Sp}$) and listener
($\mathrm{Li}$) roles; the speaker encodes its observation and
samples a candidate token sequence:
\begin{align}
  c^* \sim q(c^* \mid z^{\mathrm{Sp},v};\,
  \theta_{\mathrm{Sp}}).
  \label{eq:proposal}
\end{align}
The listener also generates a current caption $c$ from its own
decoder for the same image.
It then decides whether to accept the speaker proposal $c^*$ or
retain this listener-generated caption.
The target posterior factorizes as
$\pi(c) := p(c \mid z^{\mathrm{Sp},v}, z^{\mathrm{Li},v})
\propto p(z^{\mathrm{Li},v} \mid c)\,
p(c \mid z^{\mathrm{Sp},v})$
by conditional independence of the agents given $c$, which
follows from the CPC generative model
\citep{taniguchi2024cpc,matsui2025mhcg}, where $p(c \mid z^{\mathrm{Sp},v})$ is the speaker's posterior
over captions and $p(z^{\mathrm{Li},v} \mid c)$ is the
listener's likelihood of its visual features under caption $c$,
evaluated through the ProbVLM density
(Eq.~\ref{eq:probvlm}).
The proposal distribution is the speaker's conditional
generative distribution $q(c^* \mid z^{\mathrm{Sp},v})$.
Operationally, we use a one-step MH-style independence update in
which the listener compares the speaker proposal against its own
current caption.
Under the \emph{self-consistency approximation}
$p(c \mid z^{\mathrm{Sp},v}) \approx
q(c \mid z^{\mathrm{Sp},v})$
\citep{matsui2025mhcg}, the speaker-dependent terms in the
standard MH ratio cancel, and the acceptance probability
simplifies to a likelihood ratio from the listener's
perspective:
\begin{align}
  r \;=\; \min\!\left(1,\;
    \frac{p(\hat{h}^{\mathrm{Li},v} \mid
          \hat{\lambda}^{\mathrm{Li},t}(c^*))}
         {p(\hat{h}^{\mathrm{Li},v} \mid
          \hat{\lambda}^{\mathrm{Li},t}(c))}
  \right),
  \label{eq:mh}
\end{align}
where $\hat{h}^{\mathrm{Li},v}$ is a sample from the listener's
visual ProbVLM distribution and
$\hat{\lambda}^{\mathrm{Li},t}(c)$ denotes the listener's
text-side ProbVLM density parameters predicted from caption $c$.
Because this is not a persistent chain and the self-consistency
approximation is weak at initialization, we use the ratio in
Eq.~\ref{eq:mh} as an approximate, listener-side perceptual acceptance
rule rather than claiming exact posterior sampling;
Appendix~\ref{app:self_consistency} discusses this approximation and
its empirical diagnostics, where the finite-support Jensen--Shannon
divergence decreases from $0.660$ at epoch~0 to approximately $0.35$
at epoch~40, and the Spearman correlation between decoder and
encoder/ProbVLM log scores increases from $-0.03$ to approximately
$0.77$.

For each image and communication direction in an MH epoch, one
speaker proposal and one listener current caption are generated,
and one accept/reject decision is made.
The selected caption (the proposal if accepted, otherwise the
listener-generated caption) becomes the training target for the
listener on that image.
The opposite direction is then processed analogously.

Each agent updates its text modules using the accepted token
sequences as training targets, while its visual encoder
remains frozen throughout.
Parameter updates are interleaved with the two ordered
speaker--listener role assignments: in each MH epoch, Agent~A first serves as the speaker and
Agent~B as the listener for one pass over the dataset; the
captions selected by Agent~B's MH acceptance rule are then used
to update Agent~B.
The same procedure is then applied with roles reversed (B speaks, A listens, A is updated).
The text encoder, text decoder, projection layers, and ITM head
(collectively $\theta_n$) are updated using the selected captions
as fixed targets, minimizing BLIP-style objectives
\citep{li2022blip}:
\begin{align}
  \mathcal{L}
  \;=\;
  \mathcal{L}_{\mathrm{ITC}}
  + \mathcal{L}_{\mathrm{ITM}}
  + \mathcal{L}_{\mathrm{LM}},
  \label{eq:vlm_loss}
\end{align}
where $\mathcal{L}_{\mathrm{ITC}}$ is image-text contrastive loss,
$\mathcal{L}_{\mathrm{ITM}}$ is image-text matching loss, and
$\mathcal{L}_{\mathrm{LM}}$ is language modeling loss.
The ProbVLM adapter $\phi_n$ is updated separately using
$\mathcal{L}_{\mathrm{ProbVLM}}$ \citep{upadhyay2023probvlm},
with all other parameters frozen.
The full procedure is summarized in Algorithm~\ref{alg:mhcg}
(Appendix~\ref{app:implementation}).

\section{Experimental Setup}
\label{sec:expsetup}


\paragraph{Dataset and transforms.}
We train and evaluate on MS-COCO \citep{lin2014coco}, applying
independently sampled augmentations to each agent during
interaction so that the two agents observe distinct views of
the same image.
Training and interaction are conducted on
\texttt{train2017} approximately 118K images), and all
evaluations are performed on \texttt{val2017}
(5K images), which is unseen during interaction.
During training, the two agents observe the same underlying image
through independently sampled augmentations: Random Resized Crop
with scale $0.8$--$1.0$ and Random Horizontal Flip.
All reported validation metrics use a deterministic validation
transform without random crop or flip.

\paragraph{Visual encoder conditions.}
We compare three visual-encoder conditions, each defined by a pair
of frozen encoders: one homogeneous and two heterogeneous
(Table~\ref{tab:conditions}).
The two agents always see independently augmented views during
interaction, so view-level difference is a shared premise of all
conditions.
The conditions use the same dataset, augmentation protocol, and
interaction procedure, and differ only in the frozen visual
encoders assigned to the two agents.

We instantiate the conditions using DINOv2 visual encoders
\citep{oquab2024dinov2} and an MAE visual encoder
\citep{he2022masked}.

\begin{table}[!htbp]
  \centering
  \small
  \caption{Three visual encoder conditions.}
  \label{tab:conditions}
  \begin{tabular}{@{}lll@{}}
    \toprule
    Condition & Agent A & Agent B \\
    \midrule
    \textsc{Homo}
      & DINOv2 ViT-B/14
      & DINOv2 ViT-B/14 \\
    \textsc{Hetero1}
      & DINOv2 ViT-B/14
      & DINOv2 ViT-S/14 \\
    \textsc{Hetero2}
      & DINOv2 ViT-B/14
      & MAE ViT-B/16 \\
    \bottomrule
  \end{tabular}
\end{table}

\noindent
\textsc{Homo} pairs identical DINOv2 ViT-B/14 encoders and
is the homogeneous reference condition: during
interaction, the agents still receive independently sampled
image views, but their frozen visual encoders are the same.
\textsc{Hetero1} pairs DINOv2 ViT-B/14 with DINOv2 ViT-S/14
as the first heterogeneous encoder condition
($D^v{=}768$ vs.\ $D^v{=}384$).
\textsc{Hetero2} pairs DINOv2 ViT-B/14 with MAE ViT-B/16 as the
second heterogeneous encoder condition.
We refer to \textsc{Hetero1} as moderate heterogeneity and
\textsc{Hetero2} as stronger heterogeneity, where heterogeneity is
quantified by vision--vision RSA: the Spearman correlation between
the vectorized visual RDMs constructed from each frozen encoder's
\texttt{[CLS]} features, $\rho_s(\boldsymbol{v}^{A},\boldsymbol{v}^{B})$
(Table~\ref{tab:compact_results}).

\paragraph{Communication and acceptance-rule controls.}
We compare MHCG against a no-communication baseline (NoCom, $r \equiv 0$) and,
for \textsc{Hetero1}, four acceptance-rule ablations
(Table~\ref{tab:acceptance_controls}). 
These controls ask whether successful token-sequence formation requires the
MH likelihood-ratio filter specifically, or can be explained by proposal
exchange alone, matched acceptance rates, or the listener's BLIP-style
discriminative image--text matching ability.
Unless otherwise stated, each encoder condition is trained with
the default MHCG interaction rule, in which listener-side
Metropolis--Hastings acceptance determines whether a proposed
token sequence is adopted.
The \textsc{Hetero1} controls use the same data, encoders, model
architecture, and training horizon as the corresponding MHCG run,
differing only in the communication or acceptance rule.

\begin{table}[!htbp]
  \centering
  \small
  \caption{Communication and acceptance-rule controls.}
  \label{tab:acceptance_controls}
  \begin{tabular}{@{}ll@{}}
    \toprule
    Method & Acceptance rule \\
    \midrule
    MHCG
      & Listener-side MH acceptance using the ProbVLM likelihood ratio \\
    NoCom
      & $r \equiv 0$; never accepts proposed token sequences \\
    AllAccept
      & $r \equiv 1$; accepts every proposal \\
    Random accept
      & Matches the MHCG acceptance rate, but accepts uniformly at random \\
    ITM-based
      & Uses the listener's BLIP-style ITM head score for the candidate image--token-sequence pair \\
    \bottomrule
  \end{tabular}
\end{table}

\paragraph{Evaluation Metrics}
\label{subsec:metrics}

We evaluate whether the emergent token sequences align across agents,
how much visual information they transfer, and whether images assigned the
same shared token sequence form compact clusters in fixed visual feature
spaces.
All validation metrics are measured on \texttt{val2017}; visual-feature
prediction probes are trained on \texttt{train2017} and evaluated on
\texttt{val2017}.
Full metric definitions are provided in Appendix~\ref{app:metric_details}.
Throughout this section, $X{\to}Y$ denotes a token-sequence-source /
evaluation-space pairing: Agent~$X$ produces the token sequence, and
Agent~$Y$ supplies the representation space used for evaluation.
We use Representational Similarity Analysis (RSA), which measures correspondence between representational spaces by comparing their pairwise dissimilarity structures, encoded in Representational Dissimilarity Matrices (RDMs).
In RSA notation, $\boldsymbol{t}^{A}$ and $\boldsymbol{t}^{B}$ denote token-sequence RDMs for
Agents~A and B, whereas $\boldsymbol{v}^{A}$ and $\boldsymbol{v}^{B}$ denote visual RDMs computed
from each agent's frozen visual-encoder \texttt{[CLS]}-token output (the
global image representation pooled from the first token of the ViT
output), before the learned multimodal projection heads.

For cross-agent alignment, we report top-1 cross-modal retrieval
($K{=}10$) and inter-agent text--text RSA. Retrieval asks whether a
token sequence produced by one agent can identify the corresponding
image in the other agent's multimodal space, and text--text RSA
measures convergence between the agents' token-sequence geometries.
For visual information transfer, we use two text-to-vision measures:
$\Delta R^2$, a permutation-corrected linear prediction score from
token indicators to visual PCs, and vision--text RSA, the Spearman
correlation between token-sequence and visual RDMs. Within-agent
pairings ($A{\to}A$, $B{\to}B$) measure how much visual information each agent's token sequence carries about its own visual space, whereas
cross-agent pairings ($A{\to}B$, $B{\to}A$) measure transfer to the
other agent's visual space.
We measure the visual specificity of each shared token sequence's
image set using the global-normalized visual radius in fixed CLS
measurement spaces, distinct from the agent-specific spaces used
during training.
A token sequence is shared within a seed if at least 10 validation
images receive that sequence from both agents.
Visual specificity captures the narrowness of each shared concept's
image set and is distinct from the total amount of visual information
carried by the entire code.

\section{Results and Discussions}
\label{sec:results}
\paragraph{Encoder mismatch reduces cross-agent communicability.}
In every encoder condition, MHCG outperformed all
cross-agent metrics over NoCom, with performance decreasing as encoder mismatch increased
(Table~\ref{tab:compact_results}).
The absolute performance decreased as the visual encoders became less
similar, with \textsc{Homo} generally highest and \textsc{Hetero1}
below it, but even \textsc{Hetero2} remained above the
no-communication baseline.
Thus, emergent token sequences encode visual information that transfers
across agents.
Direction-wise and positional details are in Appendix
Table~\ref{tab:full_main_results}, ~\ref{tab:positional_entropy},
~\ref{tab:singleton_delta_r2_cross}.
\begin{table}[t]
\centering
\caption{Communication summary at epoch~40 on COCO \texttt{val2017}.
Rows report seed means $\pm$ SD.
Cross-agent columns average the $A{\to}B$ and $B{\to}A$ directions;
retrieval is top-1 accuracy with $K{=}10$.  Underlines mark MHCG
values above the corresponding NoCom row.  Full direction-wise means
and standard deviations are in Appendix Table~\ref{tab:full_main_results}.}
\label{tab:compact_results}
\small
\setlength{\tabcolsep}{2.6pt}
\resizebox{\linewidth}{!}{%
\begin{tabular}{@{}llcccccc@{}}
  \toprule
  Condition
  & Method
  &\shortstack{$\rho_s(\boldsymbol{v}^{A}, \boldsymbol{v}^{B})$$\uparrow$}
  &\shortstack{$\rho_s(\boldsymbol{t}^{A}, \boldsymbol{t}^{B})$$\uparrow$}
  &\shortstack{Cross-agent\\$\rho_s(\boldsymbol{t}, \boldsymbol{v})$ $\uparrow$}
  &\shortstack{Cross-agent\\$\Delta R^2$ $\uparrow$}
  &\multicolumn{2}{c}{\shortstack{Cross-agent retrieval\\i2t (\%) $\uparrow$\quad t2i (\%) $\uparrow$}} \\
  \midrule
  \multirow{2}{*}{\textsc{Homo}}
  & MHCG
  & \multirow{2}{*}{$1.000$}
  & $\underline{0.937}{\pm}0.021$
  & $\underline{0.162}{\pm}0.054$
  & $\underline{0.312}{\pm}0.035$
  & $\underline{92.3}{\pm}1.4$
  & $\underline{75.5}{\pm}4.6$ \\
  & NoCom
  &
  & $0.044{\pm}0.057$
  & $0.079{\pm}0.019$
  & $0.050{\pm}0.017$
  & $23.0{\pm}6.1$
  & $9.6{\pm}0.7$ \\
  \addlinespace[0.15em]
  \multirow{2}{*}{\textsc{Hetero1}}
  & MHCG
  & \multirow{2}{*}{$0.417$}
  & $\underline{0.872}{\pm}0.032$
  & $\underline{0.127}{\pm}0.039$
  & $\underline{0.286}{\pm}0.004$
  & $\underline{87.6}{\pm}2.6$
  & $\underline{62.0}{\pm}7.1$ \\
  & NoCom
  &
  & $0.042{\pm}0.012$
  & $0.053{\pm}0.025$
  & $0.039{\pm}0.015$
  & $27.0{\pm}4.3$
  & $9.8{\pm}0.5$ \\
  \addlinespace[0.15em]
  \multirow{2}{*}{\textsc{Hetero2}}
  & MHCG
  & \multirow{2}{*}{$0.074$}
  & $\underline{0.673}{\pm}0.086$
  & $\underline{0.088}{\pm}0.047$
  & $\underline{0.119}{\pm}0.073$
  & $\underline{59.0}{\pm}7.1$
  & $\underline{35.5}{\pm}12.0$ \\
  & NoCom
  &
  & $0.015{\pm}0.005$
  & $0.037{\pm}0.013$
  & $0.025{\pm}0.016$
  & $22.4{\pm}6.0$
  & $10.0{\pm}0.2$ \\
  \bottomrule
\end{tabular}%
}
\end{table}


\paragraph{Moderate heterogeneity yields fewer shared sequences but no systematic reduction in per-sequence visual specificity.}
We evaluate per-sequence visual concentration in fixed CLS
measurement spaces, distinct from the agent-specific spaces that
shaped their formation, to avoid artificially inflating specificity
scores.
We computed the global-normalized visual radius of each shared token
sequence in three fixed CLS spaces: DINOv2 ViT-B, DINOv2 ViT-S,
and MAE ViT-B.
\textsc{Hetero1} showed numerically smaller mean and median radii
than \textsc{Homo} across all three measurement spaces
(Table~\ref{tab:measurement-space-normalized-radius}), but formed
39\% fewer shared sequences.
Bootstrap CIs of the per-seed radius difference overlapped zero in
two of three seeds, so the per-sequence difference should be treated
as suggestive rather than statistically established at $n{=}3$ seeds.
Accounting for shared-sequence count, \textsc{Homo} has greater
system-level vocabulary coverage, consistent with its higher
$\Delta R^{2}$ reported above.
\begin{table}[!htbp]
  \centering
  \caption{Global-normalized visual radius in fixed CLS measurement spaces.
  Seed means $\pm$ SD over three seeds; lower = more visually specific; underlines mark row minima.}
  \label{tab:measurement-space-normalized-radius}
  \small
  \setlength{\tabcolsep}{4.5pt}
  \begin{tabular}{@{}llcc@{}}
    \toprule
    Measurement space & Condition & Mean radius $\downarrow$ & Median radius $\downarrow$ \\
    \midrule
    \multirow{3}{*}{DINOv2 ViT-B}
      & \textsc{Homo} & $0.525{\pm}0.016$ & $0.556{\pm}0.011$ \\
      & \textsc{Hetero1} & $\underline{0.505}{\pm}0.056$ & $\underline{0.499}{\pm}0.144$ \\
      & \textsc{Hetero2} & $0.654{\pm}0.060$ & $0.804{\pm}0.092$ \\
    \addlinespace
    \multirow{3}{*}{DINOv2 ViT-S}
      & \textsc{Homo} & $0.530{\pm}0.011$ & $0.539{\pm}0.016$ \\
      & \textsc{Hetero1} & $\underline{0.499}{\pm}0.042$ & $\underline{0.492}{\pm}0.128$ \\
      & \textsc{Hetero2} & $0.644{\pm}0.067$ & $0.758{\pm}0.142$ \\
    \addlinespace
    \multirow{3}{*}{MAE ViT-B}
      & \textsc{Homo} & $0.737{\pm}0.052$ & $0.699{\pm}0.026$ \\
      & \textsc{Hetero1} & $\underline{0.733}{\pm}0.010$ & $\underline{0.691}{\pm}0.008$ \\
      & \textsc{Hetero2} & $0.761{\pm}0.140$ & $0.782{\pm}0.215$ \\
    \bottomrule
\end{tabular}
\end{table}

\begin{figure}[t]
  \centering
  \includegraphics[width=\linewidth]{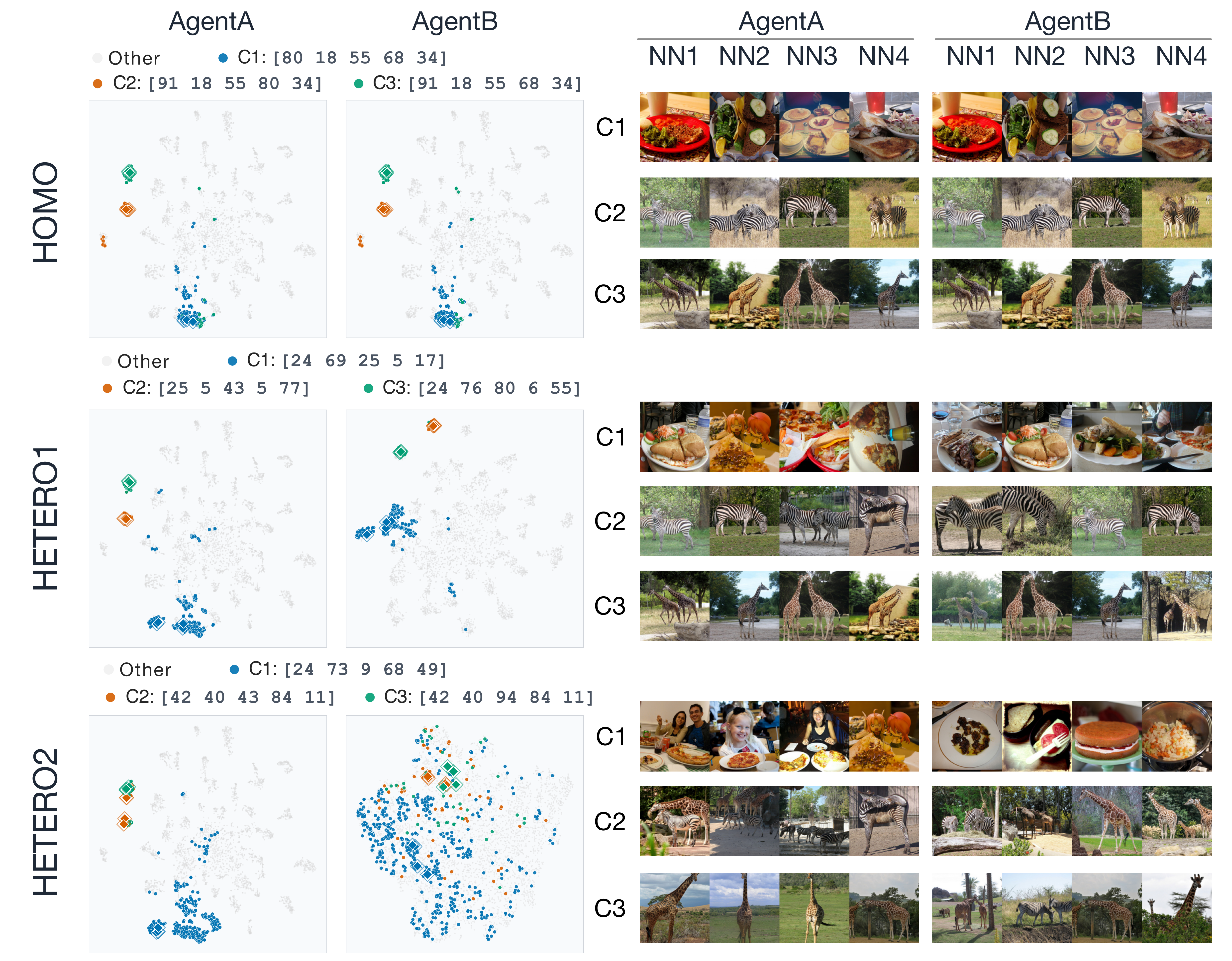}
  \caption{Matched image-set concepts across encoder conditions.
  C1--C3 are token-sequence triplets matched by image-set overlap
  after requiring both agents to assign the identical sequence to the
  same images.  The t-SNE
  panels show each agent's visual space
  colored by the matched concepts, and the image strips show nearest
  neighbors around each concept centroid.  Additional examples are in
  Appendix~\ref{app:qualitative_nn}.}
  \label{fig:qualitative}
\end{figure}

Figure~\ref{fig:qualitative} shows that all three conditions produced
visually coherent shared token sequences corresponding to recognisable
visual categories. Comparing \textsc{Homo} and \textsc{Hetero1}, the
C1--C3 image sets were largely aligned, with boundary shifts rather
than wholesale replacement (Appendix
Figure~\ref{fig:appendix_homo_hetero1_c1c3_cross_assignment}).
\textsc{Hetero2} yields far fewer shared token sequences, each
covering a larger and more object-diverse set of images
(Table~\ref{tab:token-object-resolution}), consistent with the
markedly larger visual radii reported above.


\paragraph{Representational bias toward one encoder's visual space grows
with encoder mismatch.}
Shared token sequences do not reflect both agents' visual geometries
equally: Partial-RSA bias $\Delta_X$ (Table~\ref{tab:private_structure_bias})
showed $\Delta_A > 0$ and $\Delta_B < 0$ under both heterogeneous
conditions, indicating a consistent directional tendency toward Agent~A's (DINOv2 ViT-B/14) visual geometry; however, the magnitude and behavioral consequences of this bias grow with encoder mismatch.
In \textsc{Hetero1}, where Agent~A is the higher-capacity encoder
(ViT-B vs.\ ViT-S) and their CLS-derived visual RDMs remain moderately
aligned ($\rho_s(\boldsymbol{v}^{A},\boldsymbol{v}^{B}){=}0.417$), this bias
($\Delta_A{=}0.031{\pm}0.029$) does not produce retrieval asymmetry
($A{\to}B{=}87.6{\pm}2.6\%$, $B{\to}A{=}87.5{\pm}2.6\%$ i2t).
Under \textsc{Hetero2}, where the CLS-derived visual RDMs are far more
dissimilar ($\rho_s(\boldsymbol{v}^{A},\boldsymbol{v}^{B}){=}0.074$), the bias is larger
($\Delta_A{=}0.077{\pm}0.050$) and translates into a clear retrieval
asymmetry: Agent~B's token sequences transfer more readily to
Agent~A's evaluation space than vice versa
($A{\to}B{=}47.7{\pm}13.8\%$, $B{\to}A{=}70.3{\pm}3.7\%$ i2t;
Appendix Table~\ref{tab:full_main_results}).

\begin{table}[t]
  \centering
  \caption{Private-structure bias of emergent token sequences.
  Bias is computed as
  $\Delta_X = \rho_{X,X|Y} - \rho_{X,Y|X}$; positive values indicate
  stronger dependence on Agent~$X$'s own visual geometry.}
  \label{tab:private_structure_bias}
  \small
  \begin{tabular}{@{}lcc@{}}
    \toprule
    Condition & $\Delta_A$ & $\Delta_B$ \\
    \midrule
    \textsc{Hetero1} & $0.031{\pm}0.029$ & $-0.024{\pm}0.031$ \\
    \textsc{Hetero2} & $0.077{\pm}0.050$ & $-0.037{\pm}0.021$ \\
    \bottomrule
  \end{tabular}
\end{table}


\paragraph{Listener-side MH acceptance prevents token-sequence collapse and supports visual-information transfer.}
\label{subsec:results_ablation}
Listener-side MH acceptance is critical to prevent collapse and support visual-information transfer in the tested HETERO1 condition.
We ablated the acceptance rule in the \textsc{Hetero1}
configuration, a moderately heterogeneous setting in which
the default system performs well. Table~\ref{tab:hetero1_acceptance_compact} summarizes these controls
using the same summary metrics as Table~\ref{tab:compact_results};
full direction-wise values and seed standard deviations are reported
in Appendix Table~\ref{tab:full_main_results}.  MHCG was the strongest
method in every column. The listener-side MH acceptance filter
both prevented degenerate token-sequence formation and enabled
the token sequences to carry visual information across agents.
\begin{table}[!htbp]
\centering
\caption{\textsc{Hetero1} acceptance-rule ablations at epoch~40.
Rows report seed means $\pm$ SD over available seeds.  Cross-agent columns
average $A{\to}B$ and $B{\to}A$, and the fixed vision--vision RSA
($0.417$) is omitted.  Underlines mark column-wise maxima.}
\label{tab:hetero1_acceptance_compact}
\small
\setlength{\tabcolsep}{3.4pt}
\begin{tabular}{@{}lccccc@{}}
  \toprule
  Method
  &\shortstack{$\rho_s(\boldsymbol{t}^{A}, \boldsymbol{t}^{B})$ $\uparrow$}
  &\shortstack{Cross-agent\\$\rho_s(\boldsymbol{t}, \boldsymbol{v})$ $\uparrow$}
  &\shortstack{Cross-agent\\$\Delta R^2$ $\uparrow$}
  &\multicolumn{2}{c}{\shortstack{Cross-agent retrieval\\i2t (\%) $\uparrow$\quad t2i (\%) $\uparrow$}} \\
  \midrule
  MHCG
  & $\underline{0.872}{\pm}0.032$
  & $\underline{0.127}{\pm}0.039$
  & $\underline{0.286}{\pm}0.004$
  & $\underline{87.6}{\pm}2.6$
  & $\underline{62.0}{\pm}7.1$ \\
  NoCom
  & $0.042{\pm}0.012$
  & $0.053{\pm}0.025$
  & $0.039{\pm}0.015$
  & $27.0{\pm}4.3$
  & $9.8{\pm}0.5$ \\
  AllAccept
  & $0.774{\pm}0.236$
  & $0.094{\pm}0.021$
  & $0.037{\pm}0.035$
  & $52.8{\pm}3.1$
  & $15.8{\pm}5.0$ \\
  Random accept
  & $0.011{\pm}0.023$
  & $0.016{\pm}0.017$
  & $0.009{\pm}0.004$
  & $47.8{\pm}5.2$
  & $10.0{\pm}0.0$ \\
  ITM-based
  & $0.718{\pm}0.166$
  & $0.100{\pm}0.043$
  & $0.099{\pm}0.030$
  & $67.8{\pm}9.8$
  & $38.3{\pm}8.9$ \\
  \bottomrule
\end{tabular}
\end{table}

The individual controls revealed distinct failure modes.  AllAccept
produced high inter-agent text--text RSA, but this agreement was degenerate:
compared with MHCG, it sharply reduced the number of unique sequences
($3.5{\pm}2.5$ vs.\ $90.8{\pm}34.0$) and cross-agent retrieval
(i2t/t2i: $52.8/15.8\%$ vs.\ $87.6/62.0\%$).
Random accept separated the acceptance rate from the acceptance
criterion: it accepted proposals at a similar rate to MHCG but remained
near chance in text-to-image retrieval and carried little visual
information.
ITM-based accept improved over Random accept, but remained below MHCG
across visual prediction, cross-agent vision--text RSA, and retrieval.
Together, these controls show that useful token-sequence formation
requires listener-side perceptual filtering, not merely proposal
exchange, high acceptance, or discriminative image--text matching.

\section{Conclusions, Limitations and Future Work}
\label{sec:limitations}
We asked which aspects of visual representational structure can be
preserved in shared symbols formed by agents with different private visual
representations, and whether those symbols align with both agents' representational
structures or reflect one agent's structure more strongly.
We showed that MHCG produces shared symbols, instantiated as discrete
token sequences, that carry visual information across agents and
outperform a no-communication baseline in cross-agent text--text
alignment, text--vision alignment, visual-feature prediction, and
image--text retrieval.
Across the homogeneous and heterogeneous encoder conditions, all
cross-agent metrics decreased with increasing encoder mismatch.
Moderate heterogeneity reduced the number of shared sequences while
preserving per-sequence visual specificity, whereas strong heterogeneity
yielded fewer and coarser shared token sequences.
Partial-RSA analysis further suggested a directional tendency toward one
agent's visual structure: $\Delta_A$ was positive and $\Delta_B$ negative
on average in both heterogeneous conditions, but the effect was small
under moderate mismatch ($\Delta_A{=}0.031{\pm}0.029$ in
\textsc{Hetero1}) and larger under strong mismatch
($\Delta_A{=}0.077{\pm}0.050$ in \textsc{Hetero2}), where it was
accompanied by a clear asymmetry in retrieval accuracy.

\textsc{Hetero1} formed fewer shared sequences than \textsc{Homo},
yet the per-sequence visual concentration was similar
across all three fixed measurement spaces. This suggests a possible
precision--coverage tradeoff, in which moderate
heterogeneity may preserve the visual specificity of individual shared
sequences while reducing the size of the shared vocabulary.
The lower visual-feature prediction performance under moderate
heterogeneity is consistent with
this reduced coverage, because fewer distinct visual concepts are named
and less visual information transfers across agents overall.
Whether the preserved per-sequence specificity reflects genuine sharpening
under heterogeneous constraints or a selection bottleneck that retains only
encoder-invariant concepts remains unresolved at $n{=}3$ seeds.
Future work with larger seed sets, controlled vocabulary sizes, and
extended training could decouple these explanations and determine
whether heterogeneous constraints can improve per-sequence concept
quality without sacrificing vocabulary coverage.

Several limitations bound the scope of our conclusions, but they also
point to larger questions about how perceptual and linguistic
representations co-develop through interaction.
A broader goal is to understand how private visual representations and
shared symbolic systems are jointly constructed. Our study isolates one
part of this problem by freezing the visual encoders, which 
leaves open what happens when perception itself adapts during communication.
Future work should ask whether shared symbols reshape private perceptual spaces, not only
whether fixed perceptual spaces constrain shared symbols.

A second limitation is the restricted range of empirical conditions: we tested only 
three encoder pairings on a single natural-image dataset. Results of \textsc{Hetero2} suggests that large
representational gaps can produce coarser and more asymmetric shared
token sequences, yet additional encoder families, sensory modalities,
datasets, and larger agent populations are needed to determine whether
this pattern reflects a general relation between representational heterogeneity and
symbol sharing.
Third, the vocabulary ($|\mathcal{V}|=100$) and sequence length ($L=5$)
define a small token-sequence space.
Larger vocabularies or longer sequences may
support more fine-grained concepts while making decentralized coordination
harder.

Finally, our MHCG update is a one-step MH-style approximation rather
than a persistent Markov chain, and the cancellation of
speaker-dependent terms relies on a self-consistency approximation
between the decoder proposal distribution and the speaker-side
posterior. Appendix~\ref{app:self_consistency} shows that this approximation is weakest at random initialization and
may affect the early dynamics of token-sequence formation. Future work should
investigate better calibrated acceptance ratios, asymmetric learning
rates, auxiliary alignment objectives, and more scalable variants of
decentralized inference under strong perceptual heterogeneity.



\bibliographystyle{abbrvnat}
\bibliography{references}
\clearpage
\appendix


\section{Correspondence between BLIP Training Losses and the Approximate M-step}
\label{app:mstep}

We relate the MHCG training procedure to a Monte Carlo
EM-like update for the CPC generative model, and clarify which
part of the BLIP training objective corresponds to the parameter
update after MH caption sampling.
The correspondence is approximate: the MH step supplies a sampled
caption state, while the neural update optimizes a conditional
captioning surrogate rather than the exact complete-data
likelihood of the full generative model.

The CPC generative model assumes a shared latent token sequence
$c \in \mathcal{V}^L$ that generates each agent's internal
representations:
\begin{align}
  p(c, Z^A, Z^B, o^A, o^B;\, \theta,\phi)
  &= p(c)
     \prod_{n \in \{A,B\}}
     p(Z^n \mid c;\, \theta_n,\phi_n)\, p(o^n \mid Z^n),
  \label{eq:generative}
\end{align}
where $\theta = (\theta_A, \theta_B)$ denotes the BLIP-style VLM
parameters and $\phi=(\phi_A,\phi_B)$ denotes the ProbVLM
parameters.\footnote{Following \citet{matsui2025mhcg},
we do not specify the prior $p(c)$ concretely, as it is not used
during training or inference.}
\citet{matsui2025mhcg} represent $Z^n$ as a single latent
variable.
We decompose it into three layers:
$Z^n = \{z^{n,t},\, h^n,\, z^{n,v}\}$, where $z^{n,t}$ are
text features produced by the text encoder, $z^{n,v}$ are
visual features produced by the frozen visual encoder, and
$h^n$ is a probabilistic embedding in a shared latent space
produced by ProbVLM.
The conditional factorizes as:
\begin{align}
  p(Z^n \mid c;\, \theta_n, \phi_n)
  = p(z^{n,t} \mid c;\, \theta_n)\,
    p(h^n \mid z^{n,t};\, \theta_n, \phi_n)\,
    p(z^{n,v} \mid h^n;\, \theta_n, \phi_n).
  \label{eq:generative_factorize}
\end{align}
This decomposition allows each modality to maintain its own
feature space while $h^n$ provides a probabilistic bridge
between modalities (Section~\ref{subsec:generative}).

\subsection{Approximate M-step Derivation}

In an ideal MCEM treatment, the E-step would draw caption samples
from the posterior
$p(c \mid O;\, \theta^{\mathrm{old}}, \phi^{\mathrm{old}})$
and the M-step would maximize the expected complete-data
log-likelihood. In MHCG, the MH communication step plays the role
of an approximate E-step: for each image it returns a selected
caption state $c^{(s)}$.
With one Monte Carlo sample, the exact per-agent M-step would
involve the marginal likelihood
\begin{align}
  \max_{\theta_n,\phi_n}
  \log p(o^n, c^{(s)};\, \theta_n, \phi_n)
  =
  \max_{\theta_n,\phi_n}
  \log p(o^n \mid c^{(s)};\, \theta_n, \phi_n)
  + \mathrm{const.},
  \label{eq:app_ideal_mstep}
\end{align}
where the caption prior is independent of the agent parameters.

In our three-layer generative model
(Eq.~\ref{eq:generative_factorize}), the conditional
$p(o^n \mid c^{(s)};\, \theta_n,\phi_n)$ involves marginalizing
over all intermediate latent variables:
\begin{align}
  p(o^n \mid c^{(s)};\, \theta_n,\phi_n)
  = \iiint
    p(z^{n,t} \mid c^{(s)};\, \theta_n)\,
    p(h^n \mid z^{n,t};\, \theta_n,\phi_n)\,
    p(z^{n,v} \mid h^n;\, \theta_n,\phi_n)\,
    p(o^n \mid z^{n,v})\;
    \mathrm{d}z^{n,t}\, \mathrm{d}h^n\, \mathrm{d}z^{n,v}.
  \label{eq:app_marginal}
\end{align}
This integral is intractable and is not the objective optimized by
the implemented BLIP update. Instead, because the visual encoder is
frozen, each observation can be represented by a fixed feature
$z^{n,v}=\mathrm{Enc}^{n,v}(o^n)$, and the VLM update trains the
decoder to predict the selected caption from this fixed visual
feature. We therefore use the following conditional surrogate
objective for the text module:
\begin{align}
  \theta_n^{\mathrm{new}}
  \approx
  \arg\max_{\theta_n}
  \log q_{\theta_n}(c^{(s)} \mid z^{n,v}),
  \label{eq:app_surrogate_mstep}
\end{align}
where $q_{\theta_n}$ denotes the decoder distribution
implemented by the BLIP-style text decoder.
This is a generalized or surrogate M-step: it increases the
conditional likelihood of the sampled caption under the agent's
own visual representation, but it is not claimed to be the exact
maximizer of Eq.~\ref{eq:app_ideal_mstep}.

Under the autoregressive factorization of the text decoder,
\begin{align}
  \log q_{\theta_n}(c^{(s)} \mid z^{n,v})
  = \sum_{l=1}^{L}
    \log q_{\theta_n}(c^{(s)}_l
    \mid c^{(s)}_{<l},\, z^{n,v})
  = -\mathcal{L}_{\mathrm{LM}}.
  \label{eq:app_lm_correspondence}
\end{align}

\textbf{Therefore, minimizing the language modeling loss
$\mathcal{L}_{\mathrm{LM}}$ with the selected caption
$c^{(s)}$ as target exactly optimizes the conditional captioning
surrogate in Eq.~\ref{eq:app_surrogate_mstep}.}
The three-layer latent structure
(Eq.~\ref{eq:generative_factorize}) still matters for MH
acceptance, because the listener evaluates captions through the
ProbVLM density over the shared latent space. It is not explicitly
marginalized in the BLIP decoder update.
The ProbVLM parameters $\phi_n$ are updated separately with
$\mathcal{L}_{\mathrm{ProbVLM}}$ while $\theta_n$ is frozen, so
their update is best viewed as an additional approximation that
calibrates the density model used in the next MH step.

\subsection{Role of Auxiliary Losses}

The image-text contrastive loss $\mathcal{L}_{\mathrm{ITC}}$ and
the image-text matching loss $\mathcal{L}_{\mathrm{ITM}}$ are not
part of the conditional surrogate in
Eq.~\ref{eq:app_surrogate_mstep}.
They are auxiliary objectives that maintain the representations on
which subsequent MH acceptance decisions depend:

\paragraph{$\mathcal{L}_{\mathrm{ITC}}$ (Image-Text Contrastive).}
The contrastive loss operates on the projected embeddings
$\boldsymbol{\mu}^v$ and $\boldsymbol{\mu}^t$ in the shared
latent space---the space from which the probabilistic embedding
$h^n$ (Eq.~\ref{eq:generative_factorize}) is derived via ProbVLM.
Ignoring the momentum queue and soft targets used in the BLIP
implementation, this contrastive form has the standard
InfoNCE-style interpretation as a lower bound on mutual
information \citep{oord2018representation}:
\begin{align}
  I(\boldsymbol{\mu}^v;\, \boldsymbol{\mu}^t)
  \geq \log K - \mathcal{L}_{\mathrm{ITC}},
\end{align}
where $K$ is the number of negative samples (including the memory
queue).
In practice, minimizing $\mathcal{L}_{\mathrm{ITC}}$ encourages a
shared embedding geometry in which matched image-text pairs are
close and mismatched pairs are distant.
This is critical for the ProbVLM density estimates
(Eq.~\ref{eq:probvlm}) used in the MH acceptance ratio
(Eq.~\ref{eq:mh}): if the shared embedding space of $h^n$
degrades, the density estimates become unreliable and the MH step
no longer provides useful caption samples.

\paragraph{$\mathcal{L}_{\mathrm{ITM}}$ (Image-Text Matching).}
The matching loss trains a binary classifier on the fused
cross-modal representation to predict whether an image-caption
pair is matched.
This objective encourages the cross-modal encoder to preserve
discriminative information about image-caption compatibility.
Without $\mathcal{L}_{\mathrm{ITM}}$, the $h^n$ representation
may lose its discriminative capacity, degrading the agent's
ability to evaluate the consistency of proposed captions in the
subsequent MH step.

\subsection{Summary}

The VLM training objective
$\mathcal{L} = \mathcal{L}_{\mathrm{ITC}} + \mathcal{L}_{\mathrm{ITM}} + \mathcal{L}_{\mathrm{LM}}$
can therefore be decomposed as follows:

\begin{center}
\begin{tabular}{@{}lll@{}}
  \toprule
  Loss & Role & Function \\
  \midrule
  $\mathcal{L}_{\mathrm{LM}}$ & Surrogate M-step &
    Maximize $\log q_{\theta_n}(c^{(s)} \mid z^{n,v})$ \\
  $\mathcal{L}_{\mathrm{ITC}}$ & MH support &
    Maintain $h^n$ embedding quality for ProbVLM \\
  $\mathcal{L}_{\mathrm{ITM}}$ & MH support &
    Maintain $h^n$ discriminative capacity for MH acceptance \\
  \bottomrule
\end{tabular}
\end{center}

By alternating MHCG communication and parameter updates with
$\mathcal{L}$, the agents perform an MCEM-like procedure:
the MH step updates the sampled caption state, while the neural
update fits each agent's conditional caption model and maintains
the representation and density estimates needed for future MH
rounds.

\section{Experiment Details}
\label{app:implementation}
\subsection{Self-Consistency Approximation Quality}
\label{app:self_consistency}

The MH acceptance ratio (Eq.~\ref{eq:mh}) relies on the
self-consistency approximation
$p(c \mid z^{\mathrm{Sp},v}) \approx
q(c \mid z^{\mathrm{Sp},v})$
(Section~\ref{subsec:mhcg}).
At random initialization this approximation is loose, because the
shared encoder--decoder parameters have not yet learned a
consistent mapping between captions and visual features.
As training proceeds, the LM, ITC, and ITM objectives encourage
the encoder and decoder pathways to become more compatible.
We therefore evaluate the approximation as an empirical diagnostic
rather than assuming it holds exactly.

The full token space has size $|\mathcal{V}|^L$, so exact
normalization is infeasible.
We instead compare the two sides of the approximation on a
finite candidate set.
For each validation image $o_i$, agent $n$, and checkpoint epoch,
we construct a candidate set $\mathcal{C}_i^n$ containing the
agent's greedy caption, stochastic decoder samples for the same
image, and negative captions sampled from the candidate pools of
other validation images.
All duplicate token sequences are removed.
This produces a local finite support over plausible and implausible
captions without requiring enumeration of all length-$L$ sequences.

On this support, the decoder side is measured by teacher forcing.
For a candidate caption $c=(c_1,\ldots,c_L)$ and visual feature
$z_i^{n,v}$, we compute
\begin{align}
  \ell_{\mathrm{dec}}^n(c;o_i)
  =
  \sum_{\ell=1}^{L}
  \log q_{\theta^n}
  (c_\ell \mid c_{<\ell}, z_i^{n,v}),
  \label{eq:self_consistency_decoder}
\end{align}
with padding positions ignored if present, and normalize over
$\mathcal{C}_i^n$:
\begin{align}
  \tilde{q}_i^n(c)
  =
  \frac{\exp(\ell_{\mathrm{dec}}^n(c;o_i))}
       {\sum_{c' \in \mathcal{C}_i^n}
        \exp(\ell_{\mathrm{dec}}^n(c';o_i))}.
  \label{eq:self_consistency_q}
\end{align}
The encoder/ProbVLM side is measured by asking how well the
caption-induced text density explains the same agent's visual
embedding.
Let
$\bar{h}_i^{n,v}$ denote the deterministic vision-side ProbVLM
location for image $o_i$; using the location rather than a single
sample reduces Monte Carlo noise in the diagnostic.
For each candidate caption, we compute
\begin{align}
  s_{\mathrm{enc}}^n(c;o_i)
  =
  \log p\!\left(
    \bar{h}_i^{n,v}
    \mid
    \hat{\lambda}^{n,t}(c)
  \right),
  \label{eq:self_consistency_encoder}
\end{align}
where $\hat{\lambda}^{n,t}(c)$ is obtained by passing $c$
through the text encoder, text projection, and text-side ProbVLM
adapter of the same agent.
Normalizing these scores over the same candidate set gives the
finite-support posterior proxy
\begin{align}
  \tilde{p}_i^n(c)
  =
  \frac{\exp(s_{\mathrm{enc}}^n(c;o_i))}
       {\sum_{c' \in \mathcal{C}_i^n}
        \exp(s_{\mathrm{enc}}^n(c';o_i))}.
  \label{eq:self_consistency_p}
\end{align}

We summarize the agreement between $\tilde{q}_i^n$ and
$\tilde{p}_i^n$ using four quantities: the Jensen--Shannon
divergence between the two normalized distributions, the Spearman
correlation between their unnormalized log scores over
$\mathcal{C}_i^n$, top-1 agreement of their highest-scoring
candidate, and the percentile rank assigned by
$\tilde{p}_i^n$ to the top candidate under $\tilde{q}_i^n$.
Lower Jensen--Shannon divergence, higher rank correlation, higher
top-1 agreement, and lower rank percentile indicate stronger
self-consistency.
We compute this diagnostic for the main \textsc{Hetero1} condition
(\texttt{coco\_len5\_aug\_seed0}--\texttt{seed4}) across
checkpoints, because this condition is the central heterogeneous
setting used in the main analysis.
The diagnostic is not used during training or model selection; it
only characterizes how accurately the simplified MH-style ratio
approximates the corresponding finite-support posterior comparison.

\begin{figure}[!htbp]
  \centering
  \includegraphics[width=\linewidth]{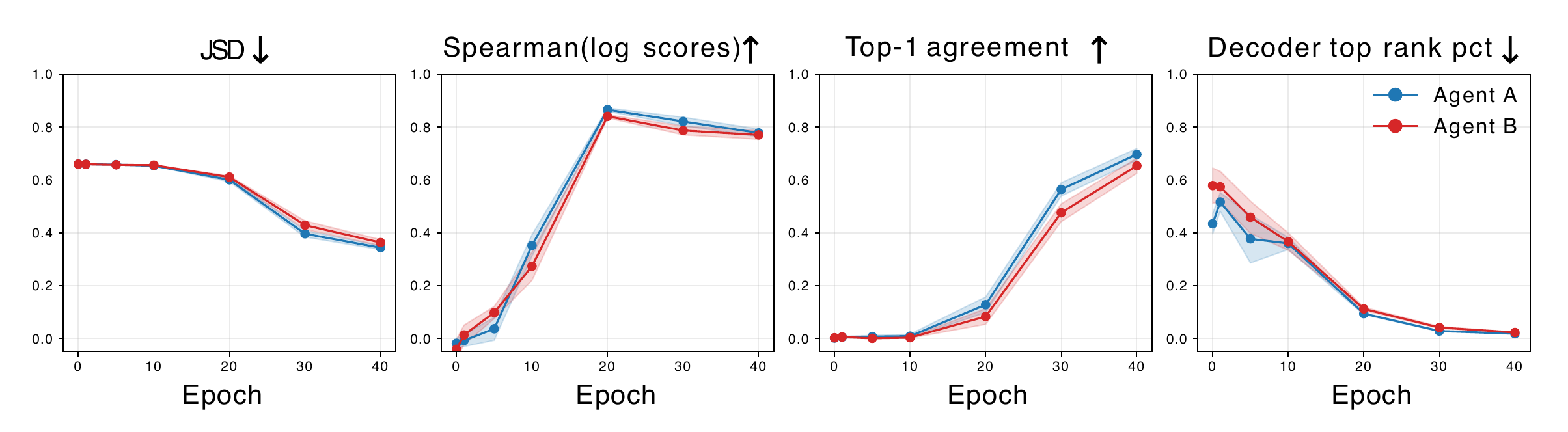}
  \caption{Self-consistency diagnostic for the main
  \textsc{Hetero1} condition
  (\texttt{coco\_len5\_aug\_seed0}--\texttt{seed4}).
  Curves show means over seeds, with shaded bands indicating the
  standard error over seeds, computed separately for Agents A and B
  on 512 validation images.
  Candidate sets contain same-image greedy and stochastic decoder
  captions plus captions sampled from other validation images, with
  duplicate token sequences removed.
  The encoder/ProbVLM posterior proxy uses the deterministic
  vision-side ProbVLM location as the visual target
  (\texttt{vision\_mu}).}
  \label{fig:self_consistency}
\end{figure}

Figure~\ref{fig:self_consistency} shows that the approximation is
poor at initialization but improves during MHCG
training.
At epoch~0, the finite-support decoder distribution and the
encoder/ProbVLM posterior proxy are nearly unrelated: the
Jensen--Shannon divergence is approximately $0.660$, the
Spearman correlation between log scores is near zero
($-0.03$ when averaged over agents), and top-1 agreement is below
$1\%$.
By epoch~40, the Jensen--Shannon divergence decreases to
approximately $0.35$, the Spearman correlation rises to
approximately $0.77$, and the top-1 agreement reaches
$65$--$70\%$ across agents.
The decoder's top-ranked caption is also assigned a much higher
rank by the encoder/ProbVLM proxy: its average rank percentile
falls from roughly the middle of the candidate set
($0.51$ at epoch~0) to the top few percent
($0.02$ at epoch~40), and it lies within the encoder/ProbVLM
top-10\% candidates in about $95\%$ of cases.

These results support the qualitative assumption used in
Eq.~\ref{eq:mh}: the tied encoder--decoder architecture does not
provide self-consistency at random initialization, but the BLIP
training objectives make the decoder-side proposal scores and the
encoder/ProbVLM-side caption scores increasingly compatible over
training.
Thus, the simplified acceptance rule should be interpreted as a
one-step MH-style approximation whose quality improves during
learning, rather than as an exact MH transition from the beginning
of training.

\begin{figure}[!htbp]
  \centering
  \includegraphics[width=0.55\linewidth]{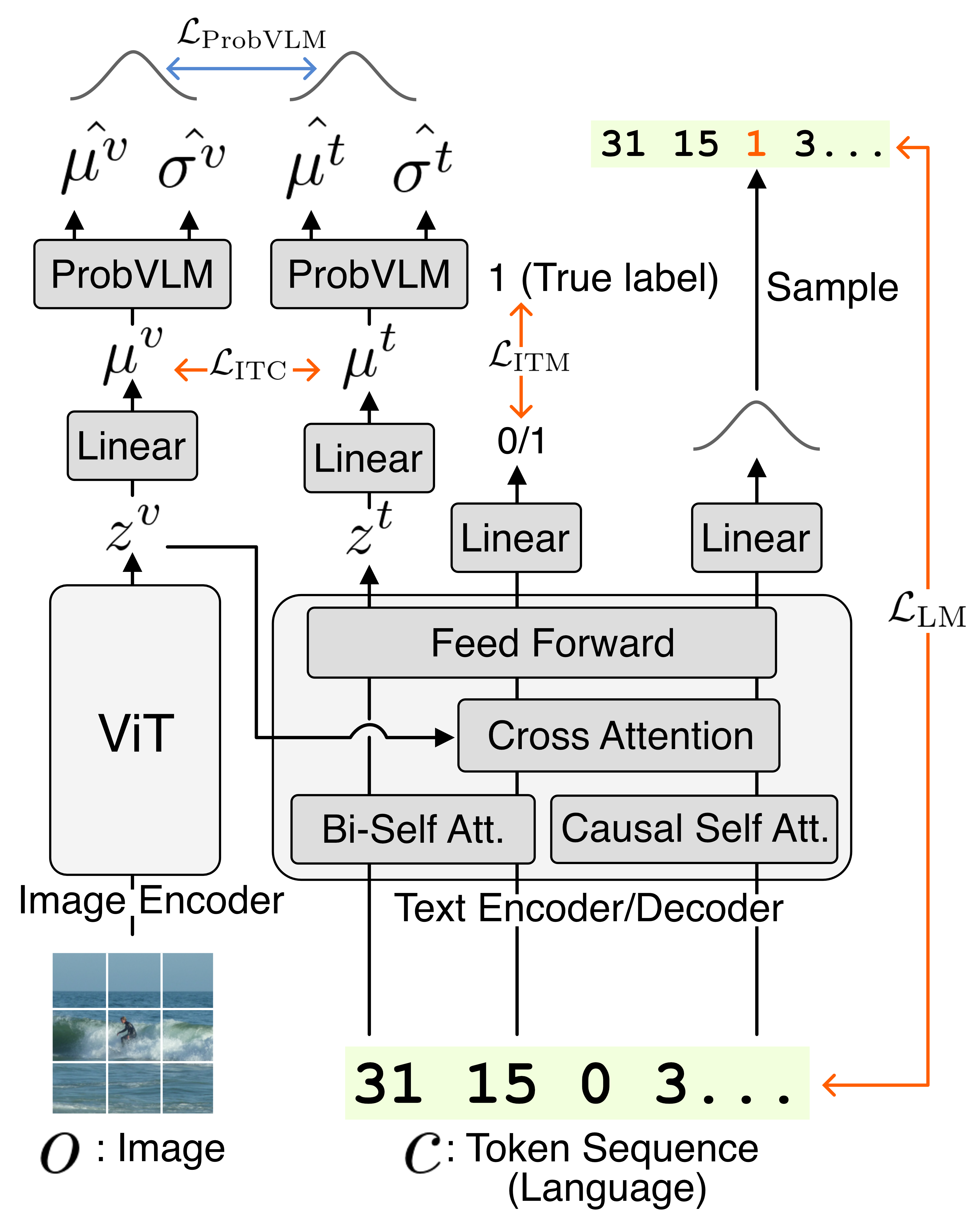}
  \caption{Single-agent BLIP-style architecture.
  A frozen ViT image encoder produces visual features $z^v$;
  a randomly initialized text encoder/decoder processes token
  sequences $c$.
  Linear projections map both modalities to a shared space
  ($\boldsymbol{\mu}^v$, $\boldsymbol{\mu}^t$), where ProbVLM
  produces density estimates used for MH acceptance.
  Training objectives: $\mathcal{L}_{\mathrm{ITC}}$ (contrastive),
  $\mathcal{L}_{\mathrm{ITM}}$ (matching),
  $\mathcal{L}_{\mathrm{LM}}$ (language modeling),
  $\mathcal{L}_{\mathrm{ProbVLM}}$ (density estimation).}
  \label{fig:architecture}
\end{figure}

\begin{algorithm}[h]
\caption{Metropolis--Hastings Captioning Game (MHCG)}
\label{alg:mhcg}
\begin{algorithmic}[1]
\Require Image dataset $\mathcal{D}$, number of epochs $E_{\mathrm{MH}}$
\State Initialize $\theta^A, \theta^B$ (text modules), $\phi^A, \phi^B$ (ProbVLM)
\For{$i = 1, \ldots, E_{\mathrm{MH}}$}
  \State $\mathcal{D}_{\mathrm{new}}^B \gets \emptyset$
  \For{each image $o_d \in \mathcal{D}$}
    \Comment{\textit{Agent A speaks to Agent B}}
    \Statex \hspace{\algorithmicindent}\hspace{\algorithmicindent}\textit{// 1.\ Perception}
    \State $z_d^{A,v} \gets \mathrm{Enc}^{A,v}(o_d)$, \; $z_d^{B,v} \gets \mathrm{Enc}^{B,v}(o_d)$ \Comment{Different augmentations}
    \State $\mu_d^{B,v} \gets \mathrm{Proj}^{B,v}(\psi(z_d^{B,v}))$
    \State $\hat{\lambda}_d^{B,v} \gets \mathrm{ProbVLM}^{B,v}(\mu_d^{B,v})$
    \State $\hat{h}_d^{B,v} \sim p(\hat{h}\mid \hat{\lambda}_d^{B,v})$
    \Statex \hspace{\algorithmicindent}\hspace{\algorithmicindent}\textit{// 2.\ Communication}
    \State $c^* \gets \mathrm{Dec}^A(z_d^{A,v})$ \Comment{Speaker proposal}
    \State $\tilde{c}_d^B \gets \mathrm{Dec}^B(z_d^{B,v})$ \Comment{Listener current caption}
    \State $c_d^B \gets \mathrm{MH\text{-}Accept}(c^*,\, \tilde{c}_d^B,\, \hat{h}_d^{B,v})$ \Comment{Eq.~\ref{eq:mh}}
    \State Add $(o_d, c_d^B)$ to $\mathcal{D}_{\mathrm{new}}^B$
  \EndFor
  \State Update $\theta^B$ on $\mathcal{D}_{\mathrm{new}}^B$ using $\mathcal{L}_{\mathrm{ITC}} + \mathcal{L}_{\mathrm{ITM}} + \mathcal{L}_{\mathrm{LM}}$
  \State Update $\phi^B$ on $\mathcal{D}_{\mathrm{new}}^B$ using $\mathcal{L}_{\mathrm{ProbVLM}}$ (with $\theta^B$ frozen)
  \State Repeat the communication and update steps with A $\leftrightarrow$ B
\EndFor
\end{algorithmic}
\end{algorithm}

\subsection{Model Specifications}
\label{app:model_specifications}

Table~\ref{tab:model_specs} summarizes the frozen visual encoders
used in each condition.
The DINOv2 encoders are loaded through the \texttt{timm}
checkpoints corresponding to the \texttt{base\_reg} and
\texttt{small\_reg} aliases
(\texttt{vit\_base\_patch14\_reg4\_dinov2.lvd142m} and
\texttt{vit\_small\_patch14\_reg4\_dinov2.lvd142m}).
The MAE encoder uses the ViT-B/16 MAE checkpoint
\texttt{mae\_pretrain\_vit\_base.pth} \citep{he2022masked}.
All visual encoders operate on $224{\times}224$ images and are
frozen throughout training.

\begin{table}[h]
  \centering
  \small
  \caption{Model specifications for the three visual encoder
  conditions.}
  \label{tab:model_specs}
  \begin{tabular}{@{}llll@{}}
    \toprule
    Condition & Agent A & Agent B & Feature dimension \\
    \midrule
    \textsc{Homo}
      & DINOv2 ViT-B/14 reg
      & DINOv2 ViT-B/14 reg
      & $768 / 768$ \\
    \textsc{Hetero1}
      & DINOv2 ViT-B/14 reg
      & DINOv2 ViT-S/14 reg
      & $768 / 384$ \\
    \textsc{Hetero2}
      & DINOv2 ViT-B/14 reg
      & MAE ViT-B/16
      & $768 / 768$ \\
    \bottomrule
  \end{tabular}
\end{table}

The text encoder/decoder uses the BERT-Tiny configuration
($L{=}2$, $H{=}128$, $A{=}2$) from the 24 smaller BERT models
released with the google-research BERT repository
\citep{turc2019wellread}: hidden dimension
$D^t = 128$, FFN intermediate
dimension 512, 2 hidden layers, 2 attention heads, max position
embeddings 16.
The projection dimension is $D^{\mathrm{proj}} = 128$.
The vocabulary size is $|\mathcal{V}| = 100$ and the
token-sequence length is $L = 5$.
The ProbVLM adapter models a diagonal generalized Gaussian
density with fixed shape parameter $\beta=2$ in the shared latent
space.
Visual encoders are frozen throughout training; the VLM-side
trainable components are the text module, projection layers, and
matching head, and the ProbVLM adapters are trained separately.
All text-module weights are initialized from scratch separately for
each run seed; no pretrained BERT checkpoint is used.
In each communication direction, the speaker proposal and the
listener current caption are both generated by the agents'
decoders for that epoch; no persistent caption table is carried
across epochs.
Each image receives one MH-style accept/reject decision per
communication direction per epoch.

\subsection{Training and Decoding Details}
\label{app:training_decoding}

\paragraph{Decoding strategy.}
During MHCG interaction (training), the speaker generates proposals
using stochastic decoding from the text decoder with nucleus
sampling ($p=0.9$), top-$k$ filtering ($k=50$), and temperature
annealing.
This stochasticity allows the one-step accept/reject update to
explore the token-sequence state space effectively.
During evaluation, we switch to deterministic greedy decoding to
assess the most representative token-sequence structure formed by the
agents.

\paragraph{Training.}
For the VLM updates, we use AdamW \citep{loshchilov2019adamw}
with $(\beta_1, \beta_2) = (0.9, 0.999)$ and weight decay $0.05$.
The VLM learning rate follows a global schedule over the configured
MHCG horizon: 10\% linear warm-up to a peak learning rate
$5 \times 10^{-5}$, followed by cosine annealing to
$1 \times 10^{-6}$; the epoch-40 update therefore uses a learning
rate of approximately $3.8 \times 10^{-5}$.
For the ProbVLM adapters, we use Adam with weight decay $1 \times
10^{-4}$ and a separate per-update cosine annealing schedule
(initial learning rate $1 \times 10^{-4}$, decaying to
$1 \times 10^{-5}$).
For each MHCG iteration, the VLM is trained for one epoch and the
ProbVLM adapter for three epochs.
The configured MHCG horizon is $E_{\mathrm{MH}} = 100$, with a batch
size of 256 per GPU; the runs reported here are stopped after 40
MHCG epochs using \texttt{--stop\_at\_epoch 40} and evaluated at the
epoch-40 checkpoint.
Following the original BLIP implementation, we employ a momentum
model and memory queues (size 131,072) for contrastive learning.
Optimizer states are re-initialized for each MH update, while the
VLM global step counter is carried across MH epochs so that the
learning-rate and soft-label schedules remain defined over the
100-epoch horizon.
The soft-label weight is linearly ramped from 0.0 to 0.4 during the
first 60\% of this horizon (60 MH epochs), reaching approximately
0.27 by the end of the reported epoch-40 update.
The caption generation temperature is annealed from 1.0 to 0.8 with
a cosine schedule over epochs 1--100, giving a temperature of
approximately 0.933 at epoch 40.

\subsection{Computational Resources}
\label{app:computational_resources}

All reported training runs were launched as single-node
distributed jobs using PyTorch DistributedDataParallel with
\texttt{torchrun --nproc\_per\_node=3}.
Each run used the three available NVIDIA A100-SXM4-80GB GPUs
on the local machine; the display GPU was not used for training.
The per-GPU batch size was 256 for MH communication, VLM updates,
and ProbVLM updates, giving an effective batch size of 768 in the
three-GPU distributed setting.
For the main experiments, each encoder condition was run for
40 MHCG epochs with three seeds (seeds 1, 2, and 4; see Appendix~\ref{app:seed_selection}) unless otherwise noted.
The \textsc{Homo}, \textsc{Hetero1}, \textsc{Hetero2}, NoCom, and
acceptance-rule ablation runs were launched sequentially by
configuration and seed.
Evaluation jobs for caption generation, retrieval, RSA, and
$\Delta R^2$ were run separately after training, typically using
one GPU per evaluation process.
Runtime estimates are based on W\&B \texttt{\_runtime} logs for
completed 40-epoch COCO runs.
Because the experiments were conducted on a shared machine where
other processes could use the same GPUs concurrently, these values
should be interpreted as observed wall-clock times rather than
controlled throughput benchmarks.
Across the completed main MHCG runs for \textsc{Homo},
\textsc{Hetero1}, and \textsc{Hetero2}, a single 40-epoch run took
9.7--23.6 hours on the three-GPU setup, with a median of
approximately 11.0 hours.
This corresponds to a nominal 29--71 A100 GPU-hours per run.

\subsection{Seed selection.}
\label{app:seed_selection}
Seed~3 is excluded from all reported results because the
\textsc{AllAccept} condition under this initialization collapsed to a
single repeated token sequence (token index~96 emitted at every
position), making cross-condition comparisons within this seed
unreliable.
For reference, the MHCG condition under \textsc{Hetero1} for seed~3
at epoch~40 yields the following retrieval and vision--text RSA values,
which are consistent with, and in most directions slightly above, the
seed-mean values reported in Table~\ref{tab:compact_results}.
Top-1 i2t retrieval (\%, $K{=}10$):
$A{\to}A = 93.1$,
$A{\to}B = 90.4$,
$B{\to}A = 90.3$,
$B{\to}B = 91.6$.
Top-1 t2i retrieval (\%, $K{=}10$):
$A{\to}A = 76.2$,
$A{\to}B = 68.8$,
$B{\to}A = 73.8$,
$B{\to}B = 69.2$.
Vision--text RSA:
$\rho_s(\boldsymbol{t}^{A}, \boldsymbol{v}^{A}) = 0.153$,
$\rho_s(\boldsymbol{t}^{A}, \boldsymbol{v}^{B}) = 0.149$,
$\rho_s(\boldsymbol{t}^{B}, \boldsymbol{v}^{A}) = 0.150$,
$\rho_s(\boldsymbol{t}^{B}, \boldsymbol{v}^{B}) = 0.154$.
The seed exclusion is therefore conservative with respect to MHCG
performance.

\subsection{Evaluation Metric Details}
\label{app:metric_details}
This appendix gives the implementation-level definitions of the
metrics summarized in Section~\ref{subsec:metrics}.
Retrieval accuracy, representational similarity, and statistics of
the image sets associated with shared token sequences are measured
on \texttt{val2017}.
For visual-feature prediction, linear probes are trained on
\texttt{train2017} and evaluated on \texttt{val2017}.
Throughout, representational similarity analysis (RSA;
\citealt{kriegeskorte2008representational}) compares
representational dissimilarity matrices (RDMs) using Spearman
correlations.

We use $X{\to}Y$ to denote a token-sequence-source/evaluation-space
pairing. Agent~$X$ supplies the token sequence, and Agent~$Y$
supplies the representation space used for evaluation. This
notation is independent of the retrieval mode and does not denote
a communication or parameter-update pass.

\paragraph{Visual specificity of shared token sequences.}
To quantify how narrowly or broadly each shared token sequence groups
images, we first identify shared token sequences within each condition
and seed.
For a token sequence $s$, the images jointly assigned by both agents
to $s$ are
\[
  \mathcal{I}_s =
  \{i : c_i^A = s \ \mathrm{and}\ c_i^B = s\},
\]
where $c_i^A$ and $c_i^B$ are the greedy validation token
sequences produced by Agents~A and B for image $i$.
Thus, \(\mathcal{I}_s\) contains the images jointly assigned by both
agents to the identical token sequence $s$.
We retain token sequences with
$|\mathcal{I}_s| \geq 10$ and call $|\mathcal{I}_s|$ the number of
shared images for sequence~$s$.
The number of retained shared token sequences measures how many
shared meanings are available in a condition.
For each shared token sequence, we compute its global-normalized visual
radius in three fixed CLS measurement spaces derived from frozen encoders:
DINOv2 ViT-B, DINOv2 ViT-S, and MAE ViT-B.
Let $r_s^Z$ be the mean cosine distance from images in
\(\mathcal{I}_s\) to their centroid in measurement space~$Z$, and
let $r_{\mathrm{global}}^Z$ be the same centroid radius over all
validation images in that measurement space.
The global-normalized visual radius in measurement space~$Z$ is
\[
  r_s^Z / r_{\mathrm{global}}^Z .
\]
Lower values mean that the token sequence selects a tighter visual
region than the validation set as a whole in that fixed measurement
space.
Within each seed, we summarize the distribution of
global-normalized visual radii over shared token
sequences by its mean and median; Table~\ref{tab:measurement-space-normalized-radius}
then reports seed means and standard deviations of these summaries.
The median shared images per sequence in
Table~\ref{tab:token-object-resolution} is the median of
$|\mathcal{I}_s|$ across shared token
sequences within a seed, again summarized across seeds.
At the object-category level, we use COCO object annotations attached
to the validation images in \(\mathcal{I}_s\).
Because a COCO image can contain multiple object categories, each
present category contributes to the category-presence distribution.
We compute its entropy $H(p_s)$ and report the effective number of
object categories, $\exp(H(p_s))$; lower values indicate narrower
object-level meanings.
We also report the top-object mass, the largest category-presence
probability in \(\mathcal{I}_s\), where higher values indicate that a
token sequence is more dominated by a single object category.
All statistics in this paragraph are computed separately for each
seed and condition before summarizing across seeds.

\paragraph{Visual-feature prediction.}
For each pairing $X{\to}Y$, we fit one-hot linear probes from
Agent~$X$'s token indicators to Agent~$Y$'s fixed visual CLS
principal-component scores.
The probes are trained on \texttt{train2017} and evaluated on
\texttt{val2017}.
PCA is fit on Agent~$Y$'s \texttt{train2017} CLS features.

For each retained PC $j$, $R^2_j(X{\to}Y)$ denotes the validation
$R^2$ of the probe trained with the true image--token pairing, and
$R^2_{j,\mathrm{perm}}(X{\to}Y)$ denotes the corresponding
validation $R^2$ after shuffling train-set token rows within the
token-sequence-source agent $X$.
We compute these values for the top 64 PCs and report the
explained-variance-weighted, permutation-corrected score:
\begin{align}
  \Delta R^2_{\mathrm{vw}}(X{\to}Y)
  =
  \sum_{j=1}^{64} w^Y_j
  \bigl(
    R^2_j(X{\to}Y)
    -
    R^2_{j,\mathrm{perm}}(X{\to}Y)
  \bigr),
\end{align}
where $w^Y_j$ is the explained-variance ratio of PC $j$ for
Agent~$Y$'s visual features, normalized over the retained PCs.
The permutation baseline is computed by repeating the train-set
shuffling procedure five times; reported values average over the
five permutations.

To characterize positional structure, we repeat the
visual-feature prediction analysis for singleton token positions and
for the full sequence.
For singleton analyses, the one-hot predictors are built from a
single token position rather than the full token sequence.
Appendix Table~\ref{tab:singleton_delta_r2_cross} reports the
cross-agent singleton-position scores for $A{\to}B$ and $B{\to}A$.

\paragraph{Representational Similarity Analysis (RSA).}
Inter-agent text--text RSA is the Spearman correlation $\rho_s$ between the
two agents' text RDMs, computed from normalized Hamming distances
between token sequences.
The Hamming distance is the fraction of mismatched token
positions, with length differences counted as mismatches.
We also track the number of unique token sequences to verify that
these communication scores are not explained by token-sequence collapse.

For each pairing $X{\to}Y$, vision--text RSA compares
Agent~$Y$'s visual RDM with Agent~$X$'s text RDM.
The token-sequence RDM is the normalized Hamming-distance RDM defined
above, and the visual RDM is the frozen-backbone CLS
cosine-distance RDM computed with the deterministic validation
transform.
The score is
\begin{align}
  \rho_s\!\left(
    \mathrm{RDM}^{X}_{\mathrm{text}},
    \mathrm{RDM}^{Y}_{\mathrm{vis}}
  \right).
\end{align}
Vision--vision RSA, $\rho_s(\boldsymbol{v}^{A}, \boldsymbol{v}^{B})$, is the Spearman
correlation between the two agents' frozen visual CLS RDMs computed
on the same validation images.
The within-agent pairings ($A{\to}A$, $B{\to}B$) measure how much
visual information each agent's own token sequence carries.
The cross-agent pairings ($A{\to}B$, $B{\to}A$) measure
inter-agent transfer.
To characterize positional structure, we repeat the
visual-feature prediction analysis for singleton token positions
and for the full sequence.
Appendix~\ref{app:positional} reports the singleton-position
analysis.

We also compute partial correlations to control the effect of visual similarity on vision--text RSA.
Let $\rho_{X,Y|Z} =
\rho_s(\mathrm{RDM}_{\mathrm{text}}^X,\,
\mathrm{RDM}_{\mathrm{vis}}^Y \mid
\mathrm{RDM}_{\mathrm{vis}}^Z)$ denote the Spearman partial
correlation between Agent~$X$'s token-sequence RDM and Agent~$Y$'s
visual RDM, controlling for Agent~$Z$'s visual RDM.
Operationally, the RDM vectors entering each partial correlation are
first converted to average ranks (ties receive their mean rank),
centered, and then combined with the standard partial-correlation
formula.
This isolates the association between the token sequence and one agent's
visual structure that is not explained by the other agent's
visual structure.
We summarize these as a \emph{private-structure bias}:
\begin{align}
  \Delta_X = \rho_{X,X|Y} - \rho_{X,Y|X},
\end{align}
where $\Delta_X \approx 0$ indicates that Agent~$X$'s token sequence
reflects structure common to both visual spaces, and
$\Delta_X > 0$ indicates that the token sequence disproportionately
reflects Agent~$X$'s private visual structure.
Results are reported in Appendix~\ref{app:partial_rsa}.

\paragraph{Cross-agent retrieval.}
For a pairing $X{\to}Y$, Agent~$X$'s validation token sequence is
encoded by Agent~$Y$'s text encoder, and the corresponding
validation image is encoded by Agent~$Y$'s visual encoder.
The projected image and text embeddings are $\ell_2$-normalized
and scored by dot product.

For each fixed pairing, we evaluate two retrieval modes.
Image-to-text retrieval (i2t) uses an image embedding as the query
and retrieves the matching token-sequence embedding.
Text-to-image retrieval (t2i) uses a token-sequence embedding as
the query and retrieves the matching image embedding.
Here, ``text'' refers to the emergent token sequence rather than
natural language.

For each query, the candidate set contains the true paired item
and $K-1$ distractors sampled without replacement.
We average over candidate-sampling seeds 200--299 and evaluate
all validation queries unless otherwise stated.
Appendix Table~\ref{tab:full_main_results} reports top-1 accuracy with
$K{=}10$ for all four pairings
($A{\to}A$, $A{\to}B$, $B{\to}A$, $B{\to}B$), with i2t and t2i
shown separately.

\paragraph{Positional entropy.}
To measure how actively each token position is used, we compute the
entropy of the empirical token distribution at each position,
separately for each agent, seed, condition, and method.
If $p_{a,\ell}(v)$ is the validation-set frequency with which
Agent~$a$ emits token $v$ at position $\ell$, the positional entropy
is
\[
  H_{a,\ell} = -\sum_v p_{a,\ell}(v)\log_2 p_{a,\ell}(v).
\]
Higher entropy indicates more diverse use of that position, while
near-zero entropy indicates an effectively fixed or inactive
position.
Appendix Table~\ref{tab:positional_entropy} reports these values in
bits.

\section{Details of Main Results}
This section expands the main-text summary results with the
seed-averaged statistics, direction-wise scores, and qualitative
examples used to interpret them.  The subsections follow the same
order as the main Results section: full communication performance,
training dynamics, visual specificity of shared token sequences,
qualitative boundary shifts, partial-RSA asymmetry, and
position-wise analyses.

\subsection{Full direction-wise communication results}
\label{app:full_main_results}

The main summary table averages over communication directions to show
the main trend.  Table~\ref{tab:full_main_results} keeps the full
direction-wise structure: within-agent directions ($A{\to}A$ and
$B{\to}B$) indicate how much visual information a token sequence carries about its own agent's visual space, whereas cross-agent directions ($A{\to}B$ and
$B{\to}A$) test whether that information transfers to the other agent's space.

\begin{table}[p]
  \caption{Direction-wise communication accuracy and visual information
  results at epoch~40 on COCO \texttt{val2017}.
  Seed means $\pm$ SD over three seeds.
  For \textsc{Hetero1}, acceptance-rule ablations are also shown.
  Visual feature prediction reports the explained-variance-weighted
  real-minus-image-permutation $\Delta R^2$ for token-to-vision
  linear prediction over the top 64 visual PCs.
  RSA reports Spearman correlations among text and vision RDMs;
  vision--vision RSA is fixed by the encoder pair and is therefore shown
  without a standard deviation. Retrieval reports top-1 i2t and t2i accuracy ($K{=}10$,
  chance $= 10\%$) as percentages for all four directions
  ($X{\to}Y$: Agent~$X$'s text is evaluated in Agent~$Y$'s multimodal space).}
  \label{tab:full_main_results}
  \centering
  \scriptsize
  \textbf{(a) Visual feature prediction}

  \vspace{0.25em}
  \setlength{\tabcolsep}{4pt}
  \resizebox{0.88\linewidth}{!}{%
  \begin{tabular}{@{}llcccc@{}}
    \toprule
    Condition & Method
    & $\Delta R^2_{A{\to}A}$
    & $\Delta R^2_{A{\to}B}$
    & $\Delta R^2_{B{\to}A}$
    & $\Delta R^2_{B{\to}B}$ \\
    \midrule
    \multirow{2}{*}{\textsc{Homo}}
      & MHCG  & $\underline{0.310}{\pm}0.034$ & $\underline{0.310}{\pm}0.034$ & $\underline{0.313}{\pm}0.035$ & $\underline{0.313}{\pm}0.035$ \\
      & NoCom & $0.057{\pm}0.015$ & $0.057{\pm}0.015$ & $0.043{\pm}0.018$ & $0.043{\pm}0.018$ \\
    \midrule
    \multirow{5}{*}{\textsc{Hetero1}}
      & MHCG  & $\underline{0.308}{\pm}0.001$ & $\underline{0.271}{\pm}0.008$ & $\underline{0.301}{\pm}0.006$ & $\underline{0.273}{\pm}0.009$ \\
      & NoCom & $0.057{\pm}0.015$ & $0.054{\pm}0.014$ & $0.024{\pm}0.017$ & $0.025{\pm}0.019$ \\
      & AllAccept & $0.037{\pm}0.036$ & $0.036{\pm}0.034$ & $0.038{\pm}0.035$ & $0.038{\pm}0.035$ \\
      & Random accept & $0.010{\pm}0.005$ & $0.010{\pm}0.005$ & $0.009{\pm}0.008$ & $0.006{\pm}0.007$ \\
      & ITM-based & $0.099{\pm}0.033$ & $0.098{\pm}0.028$ & $0.100{\pm}0.032$ & $0.102{\pm}0.028$ \\
    \midrule
    \multirow{2}{*}{\textsc{Hetero2}}
      & MHCG  & $\underline{0.162}{\pm}0.058$ & $\underline{0.126}{\pm}0.088$ & $\underline{0.111}{\pm}0.058$ & $0.127{\pm}0.106$ \\
      & NoCom & $0.057{\pm}0.015$ & $0.037{\pm}0.026$ & $0.013{\pm}0.006$ & $\underline{0.335}{\pm}0.252$ \\
    \bottomrule
  \end{tabular}%
  }

  \vspace{0.8em}
  \textbf{(b) Representational Similarity Analysis}

  \vspace{0.25em}
  \resizebox{\linewidth}{!}{%
  \begin{tabular}{@{}llcccccc@{}}
    \toprule
    Condition & Method
    & $\rho_s(\boldsymbol{t}^{A},\boldsymbol{t}^{B})$
    & $\rho_s(\boldsymbol{v}^{A},\boldsymbol{v}^{B})$
    & $\rho_s(\boldsymbol{t}^{A},\boldsymbol{v}^{A})$
    & $\rho_s(\boldsymbol{t}^{A},\boldsymbol{v}^{B})$
    & $\rho_s(\boldsymbol{t}^{B},\boldsymbol{v}^{A})$
    & $\rho_s(\boldsymbol{t}^{B},\boldsymbol{v}^{B})$ \\
    \midrule
    \multirow{2}{*}{\textsc{Homo}}
      & MHCG  & $\underline{0.937}{\pm}0.021$ & \multirow{2}{*}{$1.000$} & $\underline{0.161}{\pm}0.054$ & $\underline{0.161}{\pm}0.054$ & $\underline{0.162}{\pm}0.054$ & $\underline{0.162}{\pm}0.054$ \\
      & NoCom & $0.044{\pm}0.057$ & & $0.082{\pm}0.027$ & $0.082{\pm}0.027$ & $0.075{\pm}0.015$ & $0.075{\pm}0.015$ \\
    \midrule
    \multirow{5}{*}{\textsc{Hetero1}}
      & MHCG  & $\underline{0.872}{\pm}0.032$ & \multirow{5}{*}{$0.417$} & $\underline{0.139}{\pm}0.033$ & $\underline{0.119}{\pm}0.051$ & $\underline{0.134}{\pm}0.028$ & $\underline{0.118}{\pm}0.048$ \\
      & NoCom & $0.042{\pm}0.012$ & & $0.082{\pm}0.027$ & $0.067{\pm}0.018$ & $0.039{\pm}0.048$ & $0.041{\pm}0.041$ \\
      & AllAccept & $0.774{\pm}0.236$ & & $0.109{\pm}0.002$ & $0.090{\pm}0.050$ & $0.098{\pm}0.009$ & $0.080{\pm}0.036$ \\
      & Random accept & $0.011{\pm}0.023$ & & $0.025{\pm}0.013$ & $0.021{\pm}0.035$ & $0.012{\pm}0.025$ & $0.008{\pm}0.015$ \\
      & ITM-based & $0.718{\pm}0.166$ & & $0.106{\pm}0.038$ & $0.094{\pm}0.047$ & $0.105{\pm}0.041$ & $0.094{\pm}0.052$ \\
    \midrule
    \multirow{2}{*}{\textsc{Hetero2}}
      & MHCG  & $\underline{0.673}{\pm}0.086$ & \multirow{2}{*}{$0.074$} & $\underline{0.134}{\pm}0.016$ & $\underline{0.063}{\pm}0.056$ & $\underline{0.112}{\pm}0.039$ & $0.078{\pm}0.058$ \\
      & NoCom & $0.015{\pm}0.005$ & & $0.082{\pm}0.027$ & $0.042{\pm}0.030$ & $0.032{\pm}0.005$ & $\underline{0.298}{\pm}0.113$ \\
    \bottomrule
  \end{tabular}%
  }
  \vspace{0.8em}
  \textbf{(c) Cross-modal retrieval}

  \vspace{0.25em}
  \setlength{\tabcolsep}{1.8pt}
  \resizebox{\linewidth}{!}{%
  \begin{tabular}{@{}llcccccccc@{}}
    \toprule
    Condition & Method
    & \multicolumn{4}{c}{Top-1 i2t} & \multicolumn{4}{c}{Top-1 t2i} \\
    \cmidrule(lr){3-6} \cmidrule(lr){7-10}
    & & A$\to$A & A$\to$B & B$\to$A & B$\to$B
    & A$\to$A & A$\to$B & B$\to$A & B$\to$B \\
    \midrule
    \multirow{2}{*}{\textsc{Homo}}
      & MHCG
      & $\underline{92.4}{\pm}1.5$ & $\underline{92.3}{\pm}1.1$
      & $\underline{92.3}{\pm}1.6$ & $\underline{92.5}{\pm}1.3$
      & $\underline{75.6}{\pm}4.7$ & $\underline{75.5}{\pm}4.3$
      & $\underline{75.4}{\pm}4.8$ & $\underline{75.5}{\pm}4.5$ \\
      & NoCom
      & $48.8{\pm}7.8$ & $23.4{\pm}3.9$
      & $22.6{\pm}10.7$ & $54.9{\pm}1.2$
      & $23.0{\pm}4.3$ & $9.8{\pm}0.6$
      & $9.4{\pm}0.8$ & $21.4{\pm}3.6$ \\
    \midrule
    \multirow{5}{*}{\textsc{Hetero1}}
      & MHCG
      & $\underline{89.7}{\pm}2.8$ & $\underline{87.6}{\pm}2.6$
      & $\underline{87.5}{\pm}2.6$ & $\underline{88.7}{\pm}2.7$
      & $\underline{65.9}{\pm}7.8$ & $\underline{59.6}{\pm}6.6$
      & $\underline{64.4}{\pm}7.6$ & $\underline{60.2}{\pm}6.8$ \\
      & NoCom
      & $48.8{\pm}7.8$ & $19.8{\pm}2.6$
      & $34.3{\pm}9.5$ & $40.4{\pm}10.3$
      & $23.0{\pm}4.3$ & $9.5{\pm}0.8$
      & $10.1{\pm}0.3$ & $12.7{\pm}1.9$ \\
      & AllAccept
      & $52.6{\pm}3.1$ & $56.2{\pm}0.4$
      & $49.5{\pm}6.5$ & $52.3{\pm}5.5$
      & $16.2{\pm}5.4$ & $15.5{\pm}4.9$
      & $16.1{\pm}5.3$ & $15.5{\pm}4.9$ \\
      & Random accept
      & $33.3{\pm}14.8$ & $43.6{\pm}9.7$
      & $52.0{\pm}4.3$ & $50.9{\pm}5.4$
      & $10.0{\pm}0.0$ & $10.1{\pm}0.0$
      & $10.0{\pm}0.0$ & $10.1{\pm}0.0$ \\
      & ITM-based
      & $70.3{\pm}9.3$ & $66.9{\pm}9.9$
      & $68.6{\pm}9.6$ & $68.4{\pm}10.0$
      & $41.2{\pm}8.7$ & $35.6{\pm}9.5$
      & $40.9{\pm}8.4$ & $36.5{\pm}9.1$ \\
    \midrule
    \multirow{2}{*}{\textsc{Hetero2}}
      & MHCG
      & $\underline{78.3}{\pm}7.4$ & $\underline{47.7}{\pm}13.8$
      & $\underline{70.3}{\pm}3.7$ & $\underline{53.4}{\pm}12.1$
      & $\underline{41.6}{\pm}8.0$ & $\underline{37.8}{\pm}12.1$
      & $\underline{33.2}{\pm}12.5$ & $\underline{36.2}{\pm}13.1$ \\
      & NoCom
      & $48.8{\pm}7.8$ & $20.1{\pm}7.4$
      & $24.7{\pm}13.3$ & $32.2{\pm}1.5$
      & $23.0{\pm}4.3$ & $9.9{\pm}0.8$
      & $10.2{\pm}0.4$ & $13.6{\pm}4.9$ \\
    \bottomrule
  \end{tabular}
  }
\end{table}

Across conditions, MHCG improves the cross-agent directions over
NoCom in visual-feature prediction, RSA, and retrieval.  The
difference between within-agent and cross-agent scores becomes most
informative under heterogeneity: within-agent scores can remain
nontrivial even without successful coordination, but the cross-agent
directions reveal whether the token sequence carries visual structure
that the other agent can use.  Among the \textsc{Hetero1} acceptance
controls, MHCG is the only method that is consistently strongest
across visual prediction, vision--text RSA, and both retrieval modes.

\FloatBarrier

\subsection{Training dynamics for communication results}
\label{app:main_dynamics}

\begin{figure}[!htbp]
  \centering
  \includegraphics[width=\linewidth]{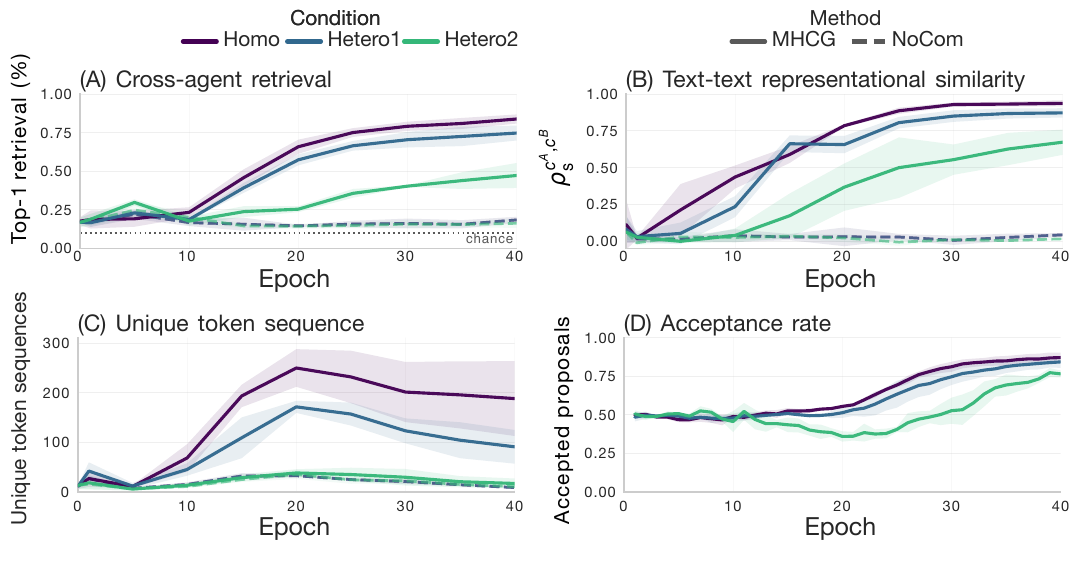}
  \caption{Token-sequence diversity, training dynamics, and token-sequence alignment over training across the
  three encoder conditions, each compared against its NoCom
  baseline.
  Colors indicate encoder condition (\textsc{Homo}, \textsc{Hetero1},
  \textsc{Hetero2}); solid curves show MHCG and dashed curves show
  NoCom.
  Shaded bands indicate mean $\pm$ standard deviation across
  available seeds.
  (A)~Cross-agent top-1 retrieval, averaged over the
  token-sequence-source/evaluation-space pairings $A{\to}B$ and
  $B{\to}A$ and over image-to-text (i2t) and text-to-image (t2i)
  retrieval modes ($K{=}10$, chance $=10\%$);
  (B)~inter-agent text--text RSA. (C)~unique token sequences, averaged over agents;
  (D)~mean acceptance rate, averaged over directions.}
  \label{fig:overview_curves}
\end{figure}

The epoch-40 tables summarize the endpoint of learning, but they do
not show whether the final system emerges through stable growth or
through transient collapse.  Figure~\ref{fig:overview_curves} therefore
plots retrieval, inter-agent token-sequence RSA, token-sequence
diversity, and MH acceptance over training for the three encoder
conditions.

\begin{figure}[!htbp]
  \centering
  \includegraphics[width=\linewidth]{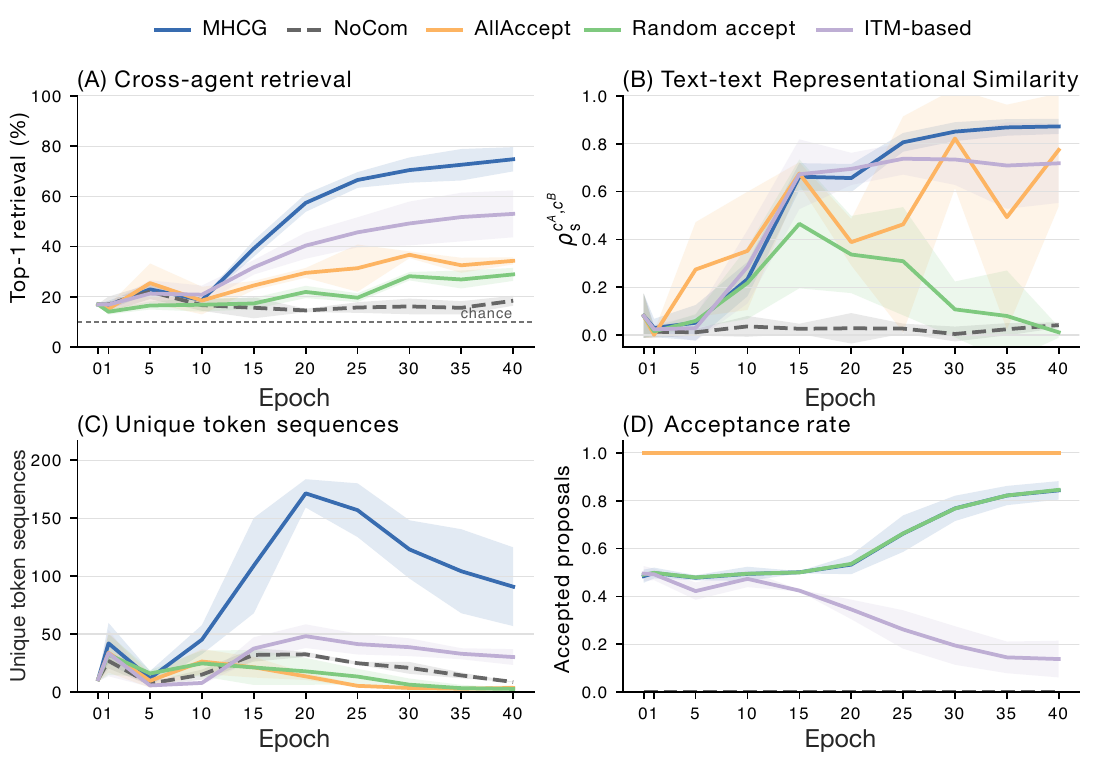}
  \caption{\textsc{Hetero1} acceptance-rule ablation dynamics.
  Curves compare MHCG, NoCom, AllAccept, Random accept, and
  ITM-based accept over training.
  Shaded bands indicate mean $\pm$ SD across three seeds.
  (A)~Cross-agent top-1 retrieval, averaged over $A{\to}B$ and
  $B{\to}A$ and over i2t/t2i retrieval modes;
  (B)~inter-agent text--text RSA based on normalized Hamming RDMs;
  (C)~unique token sequences, averaged over agents;
  (D)~mean acceptance rate, averaged over directions.}
  \label{fig:hetero1_ablation_dynamics}
\end{figure}

Figure~\ref{fig:overview_curves} shows the learning dynamics,
associated token-sequence diversity, and MH acceptance traces
corresponding to the epoch-40 retrieval values in
Appendix Table~\ref{tab:full_main_results}.
MHCG increases cross-agent retrieval and inter-agent token-sequence
alignment while maintaining a non-collapsed number of unique token
sequences.  The \textsc{Hetero2} curves remain lower and less stable,
matching the coarser endpoint statistics in the main and appendix
tables.
Figure~\ref{fig:hetero1_ablation_dynamics} provides the analogous
training dynamics for the \textsc{Hetero1} acceptance-rule controls.

The ablation dynamics clarify why listener-side MH acceptance matters.
AllAccept can produce high inter-agent token-sequence RSA, but this is
paired with sharply reduced token-sequence diversity and weak
retrieval, consistent with degenerate agreement rather than useful
communication.  Random accept shows that a similar acceptance rate is
not sufficient by itself; the acceptance decision must be tied to the
listener's perceptual evidence. ITM-based accept is intermediate,
whereas MHCG is the only setting that jointly preserves diversity and
high cross-agent retrieval.

\FloatBarrier

\subsection{Visual specificity of shared token sequences}
Table~\ref{tab:token-object-resolution} complements the fixed-space
visual-radius analysis in Table~\ref{tab:measurement-space-normalized-radius}
by asking how broad each shared token sequence is in
terms of images and COCO object categories.  The shared-sequence count
captures how many shared token sequences survive the support threshold,
whereas shared images per sequence, effective object categories, and
top-object mass characterize how broad those shared meanings are.

\begin{table}[!htbp]
  \centering
  \caption{Object-level breadth of shared token
  sequences.
  A token sequence is shared within a seed if at least 10 validation
  images receive that sequence from both agents.
  The remaining quantities are medians over shared token sequences
  within a seed, then averaged across three seeds.
  Fewer shared images per sequence and fewer effective object categories indicate
  narrower shared meanings; higher top-object mass indicates stronger
  concentration on a dominant object category.}
  \label{tab:token-object-resolution}
  \small
  \setlength{\tabcolsep}{4pt}
  \begin{tabular}{@{}lcccc@{}}
    \toprule
    Condition
    & Shared token sequences $\uparrow$
    & Shared images per sequence $\downarrow$
    & Effective object categories $\downarrow$
    & Top-object mass $\uparrow$ \\
    \midrule
    \textsc{Homo} & $77.7{\pm}18.5$ & $32.7{\pm}6.7$ & $9.5{\pm}1.0$ & $0.942{\pm}0.050$ \\
    \textsc{Hetero1} & $47.3{\pm}9.3$ & $47.3{\pm}4.2$ & $8.7{\pm}1.6$ & $0.985{\pm}0.014$ \\
    \textsc{Hetero2} & $11.0{\pm}7.2$ & $547.7{\pm}761.8$ & $20.7{\pm}9.9$ & $0.834{\pm}0.224$ \\
    \bottomrule
  \end{tabular}
\end{table}

\textsc{Hetero1} forms fewer shared token sequences than
\textsc{Homo}, but those sequences are more object-concentrated: they
have lower effective object-category breadth and higher top-object
mass.  Because this smaller sequence count could in principle bias a
direct radius comparison, Table~\ref{tab:matched_size} provides a
matched-size bootstrap control in which \textsc{Homo} is subsampled
to \textsc{Hetero1}'s token count; the matched \textsc{Homo} radii are
unchanged, confirming that the per-sequence concentration difference
is not explained by \textsc{Homo} having a long tail of highly
concentrated sequences.  The higher object concentration of
\textsc{Hetero1} sequences is therefore consistent with the
selection-bottleneck interpretation in the main text: heterogeneous
constraints retain only encoder-invariant visual concepts, resulting
in fewer but more concentrated shared sequences.  The reduced sequence
count contributes to lower system-level vocabulary coverage, consistent
with the lower $\Delta R^{2}$ under \textsc{Hetero1}.  In contrast,
\textsc{Hetero2} yields very few shared token sequences with much
more shared images per sequence and broader object coverage, consistent with
coarse token meanings under stronger visual mismatch.

\begin{table}[!htbp]
  \centering
  \caption{Matched-size bootstrap control for shared-sequence count.
  For each seed, \textsc{Homo} is subsampled to $k = \min(n_\text{Homo},\,n_\text{Hetero1})$
  tokens without replacement ($2{,}000$ iterations); \textsc{Hetero1} uses all $k$ tokens.
  The metric is the global-normalised visual radius $r/r_{\text{global}}$
  in the DINOv2~ViT-B CLS space (lower $=$ more visually specific).
  The rightmost column shows the 95\% bootstrap CI of the per-iteration
  difference $(\textsc{Homo} - \textsc{Hetero1})$.}
  \label{tab:matched_size}
  \small
  \setlength{\tabcolsep}{5pt}
  \begin{tabular}{@{}cccc c c@{}}
    \toprule
    Seed & $k$ & Condition & Original mean & Matched mean & 95\% CI of diff \\
    \midrule
    \multirow{2}{*}{1} & \multirow{2}{*}{58}
      & \textsc{Homo}    & 0.539 & 0.539 & \multirow{2}{*}{$[-0.037,\;{+}0.033]$} \\
      & & \textsc{Hetero1} & 0.542 & 0.542 & \\
    \addlinespace
    \multirow{2}{*}{2} & \multirow{2}{*}{43}
      & \textsc{Homo}    & 0.508 & 0.508 & \multirow{2}{*}{$[-0.049,\;{+}0.032]$} \\
      & & \textsc{Hetero1} & 0.517 & 0.517 & \\
    \addlinespace
    \multirow{2}{*}{4} & \multirow{2}{*}{41}
      & \textsc{Homo}    & 0.527 & 0.527 & \multirow{2}{*}{$[{+}0.033,\;{+}0.126]$} \\
      & & \textsc{Hetero1} & 0.447 & 0.447 & \\
    \bottomrule
  \end{tabular}
\end{table}

\FloatBarrier

\subsection{Additional Qualitative Examples and Boundary Shifts}
\label{app:qualitative_nn}

\begin{figure}[!htbp]
  \centering
  \begin{subfigure}{\linewidth}
    \centering
    \includegraphics[width=\linewidth]{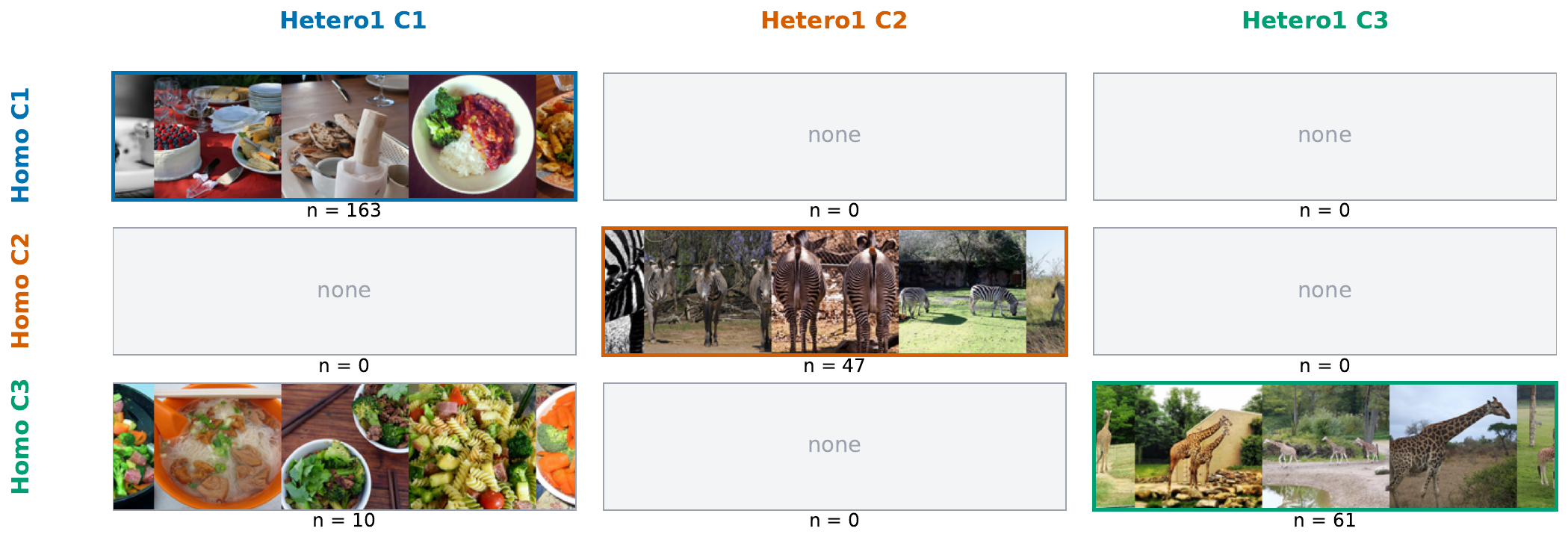}
    \caption{\textsc{Homo} vs.\ \textsc{Hetero1}}
  \end{subfigure}
  \vspace{1ex}
  \begin{subfigure}{\linewidth}
    \centering
    \includegraphics[width=\linewidth]{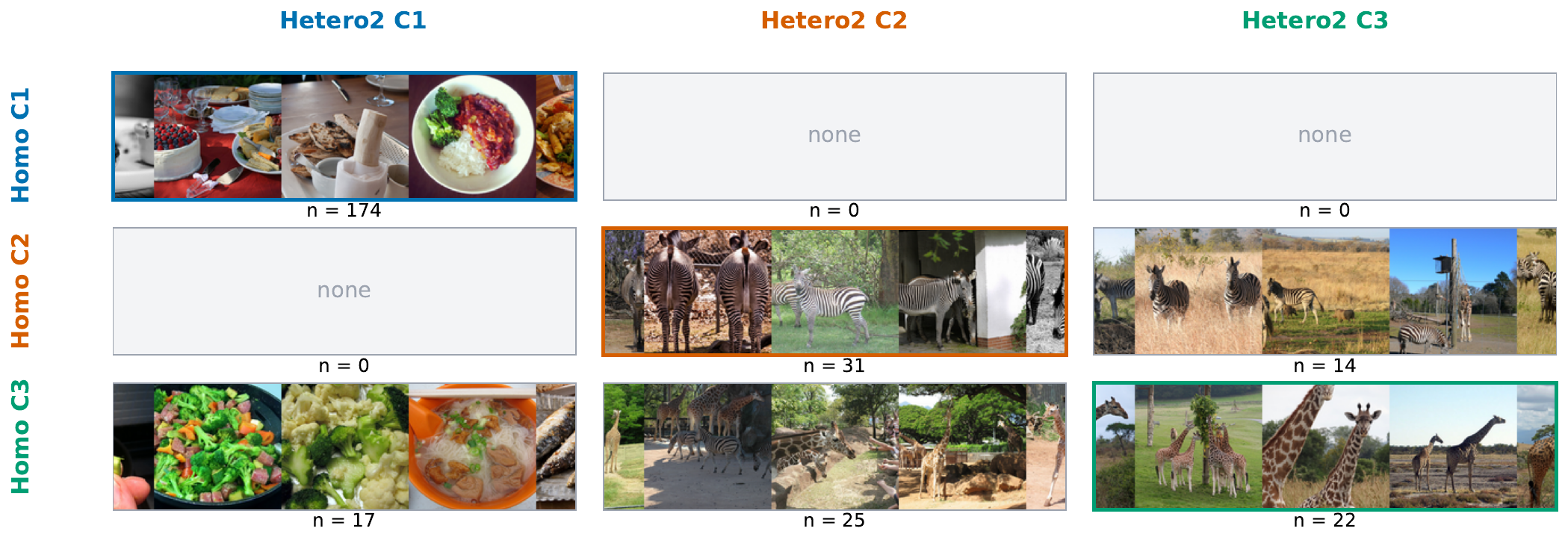}
    \caption{\textsc{Homo} vs.\ \textsc{Hetero2}}
  \end{subfigure}
  \vspace{1ex}
  \begin{subfigure}{\linewidth}
    \centering
    \includegraphics[width=\linewidth]{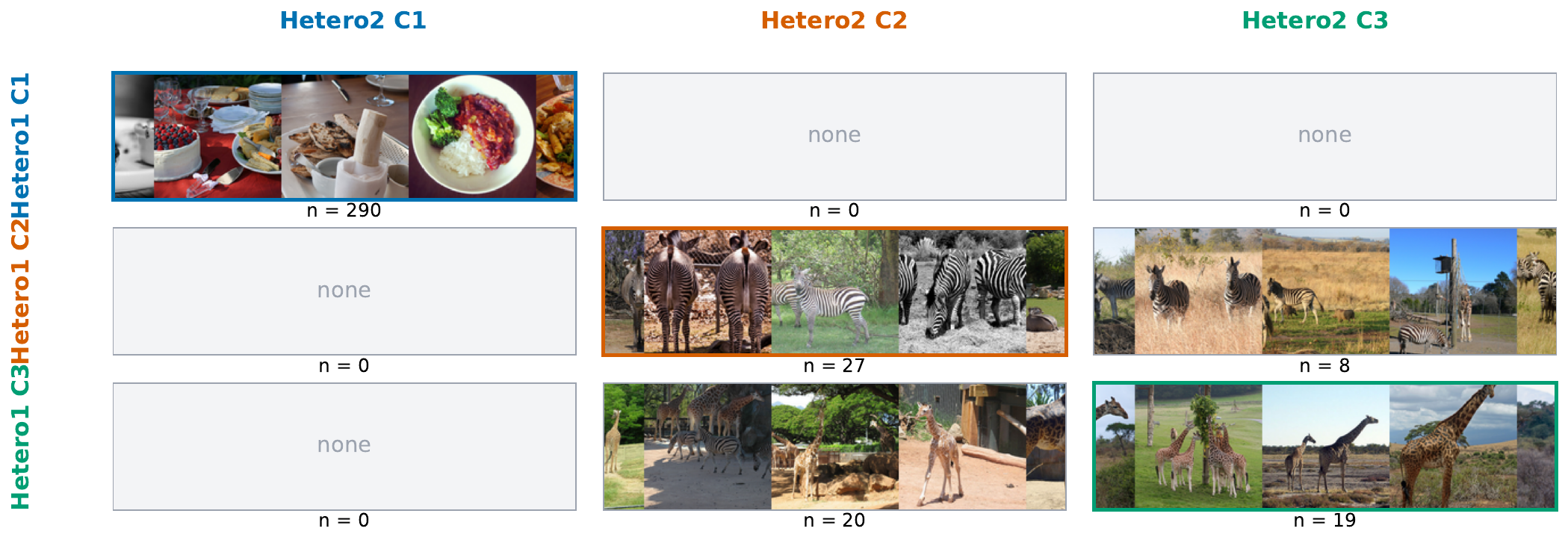}
    \caption{\textsc{Hetero1} vs.\ \textsc{Hetero2}}
  \end{subfigure}
  \caption{C1--C3 cross-assignment matrices for all condition pairs.
  Each cell shows representative images jointly assigned by both agents
  to the corresponding token-sequence concept row and column.
  Concepts are the matched C1--C3 image-set concepts used in
  Figure~\ref{fig:qualitative}; images are shown in the fixed
  DINOv2 ViT-B Agent-A measurement space.}
  \label{fig:appendix_homo_hetero1_c1c3_cross_assignment}
\end{figure}

Figure~\ref{fig:qualitative} shows representative nearest-neighbor
examples for token-sequence concepts matched across encoder
conditions by image-set overlap.
We also compare the matched \textsc{Homo} and \textsc{Hetero1}
C1--C3 sets of images jointly assigned by both agents to the
corresponding token sequence.
Figure~\ref{fig:appendix_homo_hetero1_c1c3_cross_assignment}
shows the full cross-assignment matrices for all three condition pairs.
For \textsc{Homo} vs.\ \textsc{Hetero1}, most matched concepts overlap
on the diagonal, with the only nonzero off-diagonal cell being
\textsc{Homo} C3 intersected with \textsc{Hetero1} C1 (food-like scenes).
The \textsc{Homo} vs.\ \textsc{Hetero2} and \textsc{Hetero1}
vs.\ \textsc{Hetero2} matrices show the corresponding overlap structure
under stronger encoder mismatch.

Here we provide complementary examples for the ten most frequent
shared token patterns within each encoder condition.
For each pattern, nearest neighbors are selected around the
pattern centroid in each agent's frozen CLS visual space.
When available, the centroid is computed from images for which
both agents assign that pattern to the same image; otherwise, the
candidate set falls back to images for which either agent produces
that pattern.

\begin{figure}[!htbp]
  \centering
  \includegraphics[width=\linewidth]{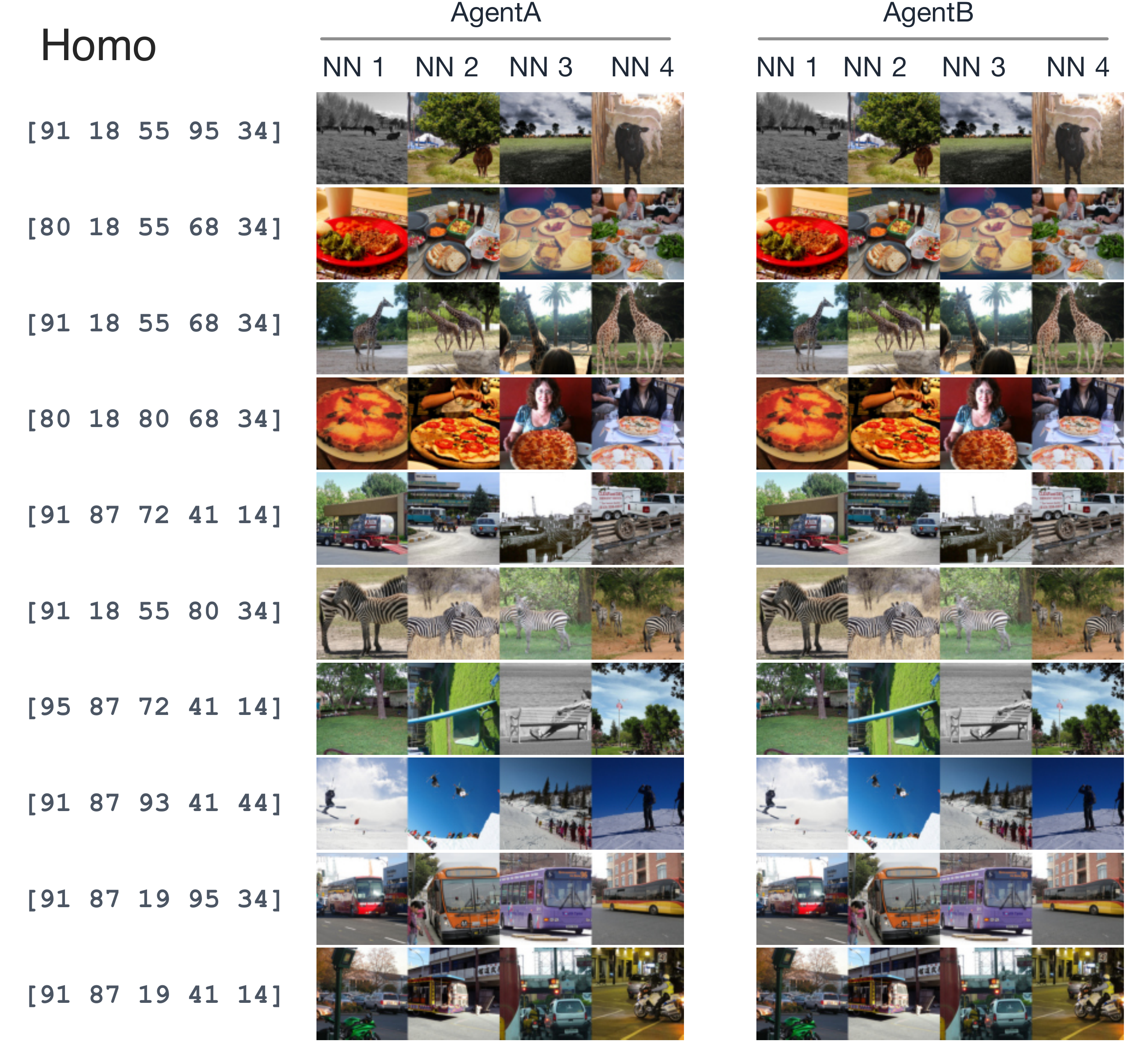}
  \caption{Additional nearest-neighbor examples for
  \textsc{Homo}.
  Rows are the ten most frequent shared token patterns.
  The left and right image strips show nearest neighbors in
  Agent~A's and Agent~B's frozen CLS spaces, respectively.}
  \label{fig:appendix_nn_view}
\end{figure}

\begin{figure}[!htbp]
  \centering
  \includegraphics[width=\linewidth]{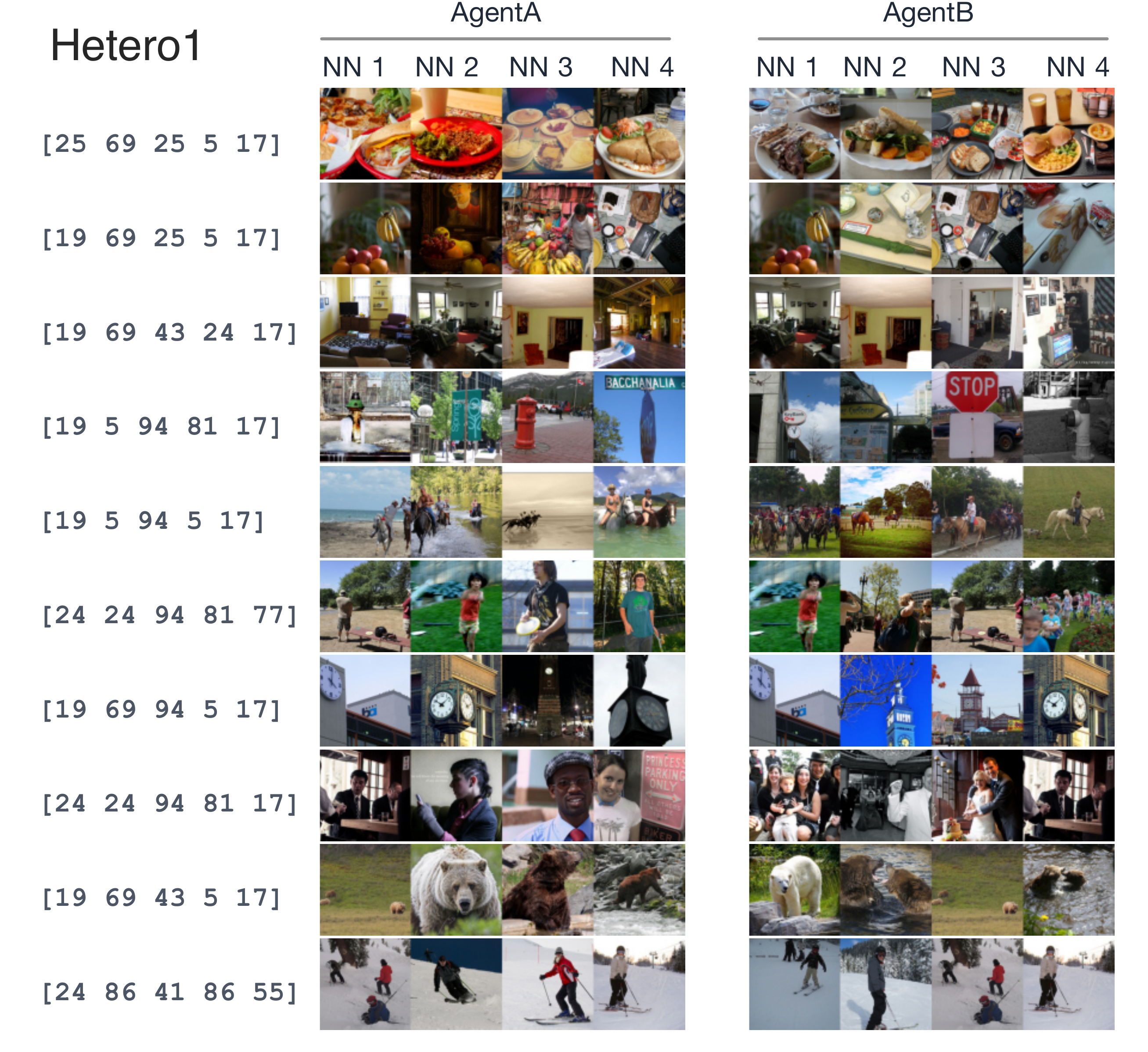}
  \caption{Additional nearest-neighbor examples for
  \textsc{Hetero1}.
  Rows are the ten most frequent shared token patterns.
  The left and right image strips show nearest neighbors in
  Agent~A's and Agent~B's frozen CLS spaces, respectively.}
  \label{fig:appendix_nn_arch}
\end{figure}

\begin{figure}[!htbp]
  \centering
  \includegraphics[width=\linewidth]{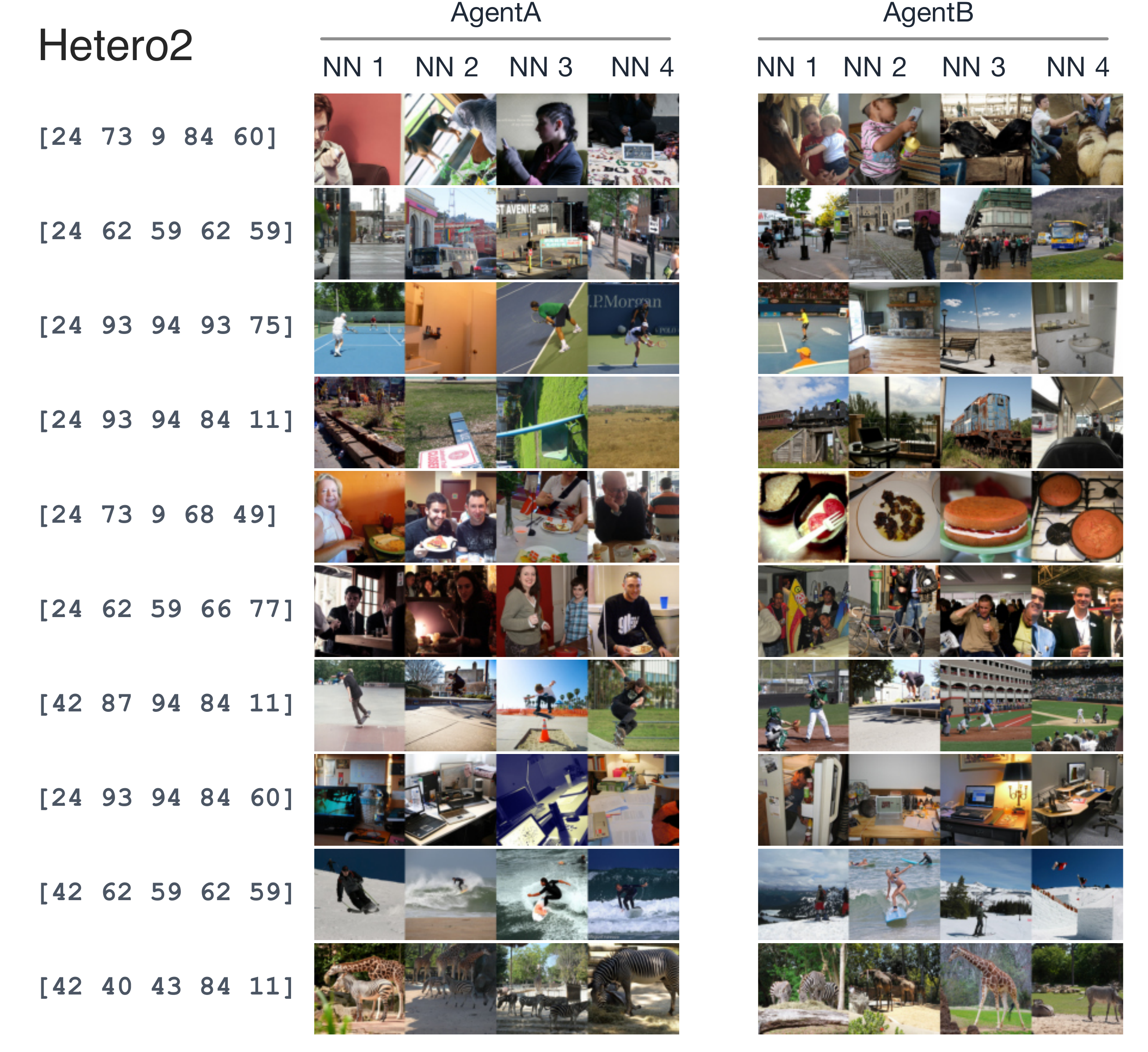}
  \caption{Additional nearest-neighbor examples for
  \textsc{Hetero2}.
  Rows are the ten most frequent shared token patterns.
  The left and right image strips show nearest neighbors in
  Agent~A's and Agent~B's frozen CLS spaces, respectively.}
  \label{fig:appendix_nn_obj}
\end{figure}

\FloatBarrier

\subsection{Partial Correlation Summary}
\label{app:partial_rsa}

Appendix Table~\ref{tab:full_main_results} reports
vision--text RSA in four directions.
Comparing within-agent values ($\boldsymbol{t}^{X}, \boldsymbol{v}^{X}$) with cross-agent
values ($\boldsymbol{t}^{X}, \boldsymbol{v}^{Y}$) provides an initial indication of whether
the token sequence reflects structure common to both agents or private to
one. Partial correlations provide a more formal decomposition by
asking whether a token-sequence RDM remains related to one visual RDM
after controlling for the other visual RDM.

Table~\ref{tab:rsa_full} reports partial Spearman correlations
at epoch~40.
Because \textsc{Homo} uses identical visual RDMs, its partial
correlations are undefined and omitted from the table. In \textsc{Hetero1},
$\Delta_A$ and $\Delta_B$ are small, consistent with the relatively
symmetric pattern observed in the main-text $\Delta R^2$ and RSA
values. Under \textsc{Hetero2}, $\Delta_A > \Delta_B$, confirming
that Agent~A's (DINOv2) token sequence disproportionately reflects
its own visual structure.
These partial-correlation results are consistent with the
within-agent vs.\ cross-agent comparisons in
Table~\ref{tab:full_main_results} and provide a complementary
statistical perspective.

\begin{table}[!htbp]
  \caption{Partial Vision--Text RSA at epoch~40 on COCO
  \texttt{val2017}.
  Means $\pm$ SD over three seeds.
  Each entry is
  $\rho_{X,Y|Z} =
  \rho_s(\mathrm{RDM}_{\mathrm{text}}^X,\,
  \mathrm{RDM}_{\mathrm{vis}}^Y \mid
  \mathrm{RDM}_{\mathrm{vis}}^Z)$, partial correlation
  between Agent~$X$'s Hamming-distance token-sequence RDM and
  Agent~$Y$'s visual RDM, controlling for Agent~$Z$'s visual RDM.
  The final columns report
  $\Delta_A = \rho_{A,A|B} - \rho_{A,B|A}$ and
  $\Delta_B = \rho_{B,B|A} - \rho_{B,A|B}$, computed per seed
  before averaging.
  \textsc{Homo} is omitted because the two visual RDMs are identical,
  making the partial correlations undefined.}
  \label{tab:rsa_full}
  \centering
  \small
  \begin{tabular}{@{}ll cccc cc@{}}
    \toprule
    Condition & Method
    & \multicolumn{4}{c}{Partial $\rho_s$}
    & \multicolumn{2}{c}{Bias} \\
    \cmidrule(lr){3-6} \cmidrule(lr){7-8}
    & & $\scriptstyle A{,}A|B$ & $\scriptstyle A{,}B|A$
    & $\scriptstyle B{,}A|B$ & $\scriptstyle B{,}B|A$
    & $\scriptstyle \Delta_A$ & $\scriptstyle \Delta_B$ \\
    \midrule
    \multirow{2}{*}{\textsc{Hetero1}}
      & MHCG
      & $0.099{\pm}0.014$ & $0.068{\pm}0.041$
      & $0.093{\pm}0.010$ & $0.070{\pm}0.040$
      & $0.031{\pm}0.029$ & $-0.024{\pm}0.031$ \\
      & NoCom
      & $0.059{\pm}0.022$ & $0.037{\pm}0.008$
      & $0.024{\pm}0.035$ & $0.028{\pm}0.023$
      & $0.022{\pm}0.016$ & $0.004{\pm}0.013$ \\
    \midrule
    \multirow{2}{*}{\textsc{Hetero2}}
      & MHCG
      & $0.130{\pm}0.013$ & $0.053{\pm}0.056$
      & $0.107{\pm}0.035$ & $0.070{\pm}0.056$
      & $0.077{\pm}0.050$ & $-0.037{\pm}0.021$ \\
      & NoCom
      & $0.079{\pm}0.026$ & $0.036{\pm}0.028$
      & $0.010{\pm}0.011$ & $0.297{\pm}0.114$
      & $0.043{\pm}0.010$ & $0.287{\pm}0.124$ \\
    \bottomrule
  \end{tabular}
\end{table}

The normalized Hamming RDM contains many tied entries, with tie density
ranging from $0.19$ (\textsc{Hetero1} MHCG) to $0.66$
(\textsc{Hetero1} NoCom seed~2). Spearman partial correlations on
heavily tied RDMs may be sensitive to the mid-rank convention. We
therefore report a Kendall $\tau_b$-based partial-RSA analogue as a
tie-aware robustness check, computed by applying the standard
partial-correlation algebra
$\tau_{X,Y|Z} = (\tau_{XY} - \tau_{XZ}\tau_{YZ}) /
\sqrt{(1-\tau_{XZ}^2)(1-\tau_{YZ}^2)}$, where each $\tau$ is the
corresponding pairwise Kendall $\tau_b$ value.

Table~\ref{tab:partial_rsa_kendall} compares Spearman and Kendall
$\tau_b$ partial-RSA bias values. The two metrics differ in absolute
magnitude (Kendall values are approximately $0.6$--$0.8$ times the
corresponding Spearman values, consistent with the standard
relationship between the two correlation measures), but agree in sign
and in the qualitative pattern across conditions. The MHCG-versus-NoCom
shift in $\Delta_B$ under \textsc{Hetero2} is preserved. The
$\Delta_B$ sign reversal between NoCom and MHCG is observed in all
three seeds under both metrics.
\begin{table}[!htbp]
  \caption{Partial-RSA under Kendall $\tau_b$.
  Spearman partial correlations are reproduced from
  Table~\ref{tab:rsa_full}; Kendall $\tau_b$-based partial scores are
  computed by applying the partial-correlation formula to pairwise
  Kendall $\tau_b$ values.}
  \label{tab:partial_rsa_kendall}
  \centering
  \small
  \setlength{\tabcolsep}{4pt}
  \begin{tabular}{@{}ll cc cc@{}}
    \toprule
    & & \multicolumn{2}{c}{Spearman partial}
      & \multicolumn{2}{c}{Kendall $\tau_b$-based partial} \\
    \cmidrule(lr){3-4} \cmidrule(lr){5-6}
    Condition & Method & $\Delta_A$ & $\Delta_B$ & $\Delta_A$ & $\Delta_B$ \\
    \midrule
    \multirow{2}{*}{\textsc{Hetero1}}
      & MHCG  & $+0.031{\pm}0.029$ & $-0.024{\pm}0.031$
              & $+0.020{\pm}0.018$ & $-0.015{\pm}0.020$ \\
      & NoCom & $+0.022{\pm}0.016$ & $+0.004{\pm}0.013$
              & $+0.014{\pm}0.010$ & $+0.003{\pm}0.009$ \\
    \midrule
    \multirow{2}{*}{\textsc{Hetero2}}
      & MHCG  & $+0.077{\pm}0.050$ & $-0.037{\pm}0.021$
              & $+0.056{\pm}0.036$ & $-0.028{\pm}0.017$ \\
      & NoCom & $+0.043{\pm}0.010$ & $+0.287{\pm}0.124$
              & $+0.031{\pm}0.008$ & $+0.212{\pm}0.090$ \\
    \bottomrule
  \end{tabular}
\end{table}
\FloatBarrier

\subsection{Positional Analyses}

\label{app:positional}

The previous analyses treat each length-5 token sequence as a single
symbolic unit.  The positional analyses ask whether the communication
system distributes information across positions or relies on a small
subset of positions.  Positional entropy measures how actively each
position varies, and singleton-position $\Delta R^2$ measures how much
cross-agent visual information can be decoded from one position at a
time.

\begin{table}[!htbp]
  \caption{Positional entropy of generated token sequences.
  Mean $\pm$ SD in bits across three seeds at epoch~40.}
  \label{tab:positional_entropy}
  \centering
  \scriptsize
  \setlength{\tabcolsep}{3pt}
  \begin{tabular}{@{}lllccccc@{}}
    \toprule
    Condition & Method & Agent
      & Pos.~1 & Pos.~2 & Pos.~3 & Pos.~4 & Pos.~5 \\
    \midrule
    \multirow{4}{*}{\textsc{Homo}} & \multirow{2}{*}{MHCG} & A
      & $2.25 \pm 0.30$ & $2.15 \pm 0.60$
      & $1.99 \pm 0.33$ & $1.78 \pm 0.52$
      & $2.01 \pm 0.60$ \\
    & & B
      & $2.25 \pm 0.29$ & $2.18 \pm 0.57$
      & $2.00 \pm 0.32$ & $1.79 \pm 0.54$
      & $2.01 \pm 0.57$ \\
    & \multirow{2}{*}{NoCom} & A
      & $1.21 \pm 0.30$ & $1.26 \pm 0.64$
      & $0.97 \pm 0.51$ & $0.95 \pm 0.12$
      & $0.82 \pm 0.72$ \\
    & & B
      & $0.91 \pm 0.39$ & $0.85 \pm 0.16$
      & $0.76 \pm 0.78$ & $1.21 \pm 0.35$
      & $0.64 \pm 0.56$ \\
    \midrule
    \multirow{4}{*}{\textsc{Hetero1}} & \multirow{2}{*}{MHCG} & A
      & $1.71 \pm 0.22$ & $2.56 \pm 0.11$
      & $1.95 \pm 0.46$ & $2.04 \pm 0.71$
      & $1.51 \pm 0.26$ \\
    & & B
      & $1.72 \pm 0.25$ & $2.56 \pm 0.11$
      & $1.97 \pm 0.46$ & $2.02 \pm 0.71$
      & $1.52 \pm 0.24$ \\
    & \multirow{2}{*}{NoCom} & A
      & $1.21 \pm 0.30$ & $1.26 \pm 0.64$
      & $0.97 \pm 0.51$ & $0.95 \pm 0.12$
      & $0.82 \pm 0.72$ \\
    & & B
      & $0.09 \pm 0.07$ & $0.30 \pm 0.30$
      & $0.07 \pm 0.06$ & $0.36 \pm 0.56$
      & $0.70 \pm 0.45$ \\
    \midrule
    \multirow{4}{*}{\textsc{Hetero2}} & \multirow{2}{*}{MHCG} & A
      & $0.21 \pm 0.29$ & $1.33 \pm 0.95$
      & $1.69 \pm 0.52$ & $1.67 \pm 0.50$
      & $1.62 \pm 0.78$ \\
    & & B
      & $0.21 \pm 0.35$ & $1.28 \pm 1.02$
      & $1.58 \pm 0.54$ & $1.60 \pm 0.54$
      & $1.27 \pm 1.16$ \\
    & \multirow{2}{*}{NoCom} & A
      & $1.21 \pm 0.30$ & $1.26 \pm 0.64$
      & $0.97 \pm 0.51$ & $0.95 \pm 0.12$
      & $0.82 \pm 0.72$ \\
    & & B
      & $0.49 \pm 0.39$ & $0.54 \pm 0.52$
      & $0.97 \pm 0.15$ & $0.89 \pm 0.68$
      & $1.00 \pm 0.46$ \\
    \bottomrule
  \end{tabular}
\end{table}

\begin{table}[!htbp]
  \caption{Cross-agent singleton-position visual-feature prediction.
  Values are explained-variance-weighted $\Delta R^2$ for singleton
  token positions at epoch~40, reported as mean $\pm$ standard
  deviation across three seeds.
  The $A{\to}B$ direction predicts Agent~B's visual PCs from Agent~A's
  singleton token position, and $B{\to}A$ is defined analogously.}
  \label{tab:singleton_delta_r2_cross}
  \centering
  \scriptsize
  \setlength{\tabcolsep}{3pt}
  \begin{tabular}{@{}lllccccc@{}}
    \toprule
    Condition & Method & Direction
      & Pos.~1 & Pos.~2 & Pos.~3 & Pos.~4 & Pos.~5 \\
    \midrule
    \multirow{4}{*}{\textsc{Homo}} & \multirow{2}{*}{MHCG} & $A{\to}B$
      & $0.111{\pm}0.028$ & $0.084{\pm}0.028$
      & $0.097{\pm}0.024$ & $0.099{\pm}0.029$
      & $0.096{\pm}0.021$ \\
    & & $B{\to}A$
      & $0.110{\pm}0.024$ & $0.085{\pm}0.026$
      & $0.099{\pm}0.024$ & $0.100{\pm}0.030$
      & $0.096{\pm}0.021$ \\
    & \multirow{2}{*}{NoCom} & $A{\to}B$
      & $0.029{\pm}0.020$ & $0.030{\pm}0.015$
      & $0.016{\pm}0.002$ & $0.010{\pm}0.006$
      & $0.015{\pm}0.013$ \\
    & & $B{\to}A$
      & $0.021{\pm}0.009$ & $0.019{\pm}0.003$
      & $0.018{\pm}0.018$ & $0.023{\pm}0.007$
      & $0.011{\pm}0.010$ \\
    \midrule
    \multirow{4}{*}{\textsc{Hetero1}} & \multirow{2}{*}{MHCG} & $A{\to}B$
      & $0.080{\pm}0.036$ & $0.130{\pm}0.015$
      & $0.094{\pm}0.028$ & $0.082{\pm}0.035$
      & $0.071{\pm}0.011$ \\
    & & $B{\to}A$
      & $0.092{\pm}0.048$ & $0.141{\pm}0.027$
      & $0.107{\pm}0.031$ & $0.085{\pm}0.030$
      & $0.078{\pm}0.017$ \\
    & \multirow{2}{*}{NoCom} & $A{\to}B$
      & $0.028{\pm}0.018$ & $0.029{\pm}0.015$
      & $0.016{\pm}0.002$ & $0.010{\pm}0.005$
      & $0.014{\pm}0.012$ \\
    & & $B{\to}A$
      & $0.004{\pm}0.003$ & $0.008{\pm}0.004$
      & $0.002{\pm}0.003$ & $0.009{\pm}0.014$
      & $0.019{\pm}0.021$ \\
    \midrule
    \multirow{4}{*}{\textsc{Hetero2}} & \multirow{2}{*}{MHCG} & $A{\to}B$
      & $0.008{\pm}0.012$ & $0.081{\pm}0.080$
      & $0.084{\pm}0.076$ & $0.068{\pm}0.053$
      & $0.072{\pm}0.043$ \\
    & & $B{\to}A$
      & $0.011{\pm}0.020$ & $0.062{\pm}0.033$
      & $0.059{\pm}0.044$ & $0.061{\pm}0.048$
      & $0.056{\pm}0.030$ \\
    & \multirow{2}{*}{NoCom} & $A{\to}B$
      & $0.019{\pm}0.014$ & $0.030{\pm}0.027$
      & $0.015{\pm}0.017$ & $0.013{\pm}0.020$
      & $0.017{\pm}0.018$ \\
    & & $B{\to}A$
      & $0.003{\pm}0.001$ & $0.003{\pm}0.003$
      & $0.006{\pm}0.004$ & $0.005{\pm}0.004$
      & $0.006{\pm}0.001$ \\
    \bottomrule
  \end{tabular}
\end{table}

Tables~\ref{tab:positional_entropy} and
\ref{tab:singleton_delta_r2_cross} report how actively each token
position is used and how much cross-agent visual information each
singleton position carries for MHCG and the NoCom baseline.
Under \textsc{Hetero1}, both agents use all positions with comparable
entropy, and singleton positions, especially positions 2--3, carry
cross-agent visual information in both directions.
Under \textsc{Hetero2}, positional entropy is more uneven and
cross-agent singleton $\Delta R^2$ is weaker and more directionally
asymmetric.
This compression of position-wise cross-agent visual information is
consistent with the finding that, under strong heterogeneity, each
shared token sequence tends to cover broader image sets.

\FloatBarrier


\section{Broader Impacts}
\label{app:broader_impacts}

This work is foundational research on emergent communication and
does not target a specific deployment. We identify the following
potential societal impacts.

\paragraph{Positive impacts.}
Understanding how agents with heterogeneous private perceptual systems
can develop shared communication through decentralized interaction
informs the design of distributed multi-agent AI systems that must
coordinate across diverse architectures without centralised control.
The decentralized update rule, which requires no parameter sharing,
cross-agent gradients, or shared reward signal, may contribute to
privacy-preserving multi-agent learning, where agents coordinate
without exposing their internal representations to one another.
At a scientific level, this work contributes to understanding how
shared symbolic systems can arise from diverse sensory experiences,
with potential relevance to cognitive science and the study of
language grounding.

\paragraph{Negative impacts and mitigations.}
The primary concern identified by our own results is
\emph{representational bias propagation}: emergent shared
representations systematically inherit one agent's visual geometry
rather than reflecting a neutral average of both.
If systems of this kind were scaled or deployed, encoder-specific
perceptual biases, including those arising from training data
or architectural choices, could be embedded in a shared language
space in ways that are difficult to detect or audit.
The representational bias analysis introduced in this work
(Section~\ref{sec:results}, Table~\ref{tab:private_structure_bias})
provides a diagnostic framework for quantifying such asymmetries,
which could be applied to audit bias in deployed emergent
communication systems.
A second, more distal concern is that autonomous inter-agent
communication protocols developed through decentralised interaction
could, in more capable settings, become difficult for humans to
interpret or monitor; the small vocabulary ($|\mathcal{V}|=100$)
and sequence length ($L=5$) used here are far from such a regime.
We do not foresee a direct path from the present work to harmful
applications, but we consider transparency about these considerations
appropriate.

\clearpage
\newpage

\end{document}